\documentclass{article}

\usepackage{microtype}
\usepackage{graphicx}
\usepackage{subcaption}
\usepackage{booktabs} 

\usepackage{hyperref}

\usepackage[preprint]{icml2026}

\usepackage{amsmath}
\usepackage{amssymb}
\usepackage{mathtools}
\usepackage{amsthm}

\usepackage[capitalize,noabbrev]{cleveref}

\usepackage[normalem]{ulem}
\usepackage{booktabs}

\usepackage{adjustbox}
\usepackage[table]{xcolor}      
\usepackage{wrapfig}
\usepackage{subcaption}
\usepackage{tabularx}
\usepackage{multirow}
\usepackage{placeins}

\theoremstyle{plain}

\theoremstyle{definition}

\theoremstyle{remark}

\usepackage[textsize=tiny]{todonotes}

\icmltitlerunning{NoiseSDF2NoiseSDF: Learning Clean Neural Fields from Noisy Supervision}

\begin{document}

\twocolumn[
  \icmltitle{NoiseSDF2NoiseSDF: Learning Clean Neural Fields from Noisy Supervision}



  \icmlsetsymbol{equal}{*}

  \begin{icmlauthorlist}
    \icmlauthor{Tengkai Wang}{equal,anu}
    \icmlauthor{Weihao Li}{equal,anu}
    \icmlauthor{Ruikai Cui}{anu}
    \icmlauthor{Shi Qiu}{cuhk}
    \icmlauthor{Nick Barnes}{anu}
  \end{icmlauthorlist}

  \icmlaffiliation{anu}{Australian National University, Canberra, ACT, Australia}
  \icmlaffiliation{cuhk}{The Chinese University of Hong Kong, Hong Kong SAR, China}

  \icmlcorrespondingauthor{Weihao Li}{weihao.li1@anu.edu.au}
  \icmlcorrespondingauthor{Shi Qiu}
  {shiqiu@cse.cuhk.edu.hk}
  \icmlcorrespondingauthor{Tengkai Wang}{Tengkai.Wang@anu.edu.au}

  \icmlkeywords{Noise2Noise, Noisy Supervision, Self-supervised Denoising, Neural Fields, Signed Distance Functions, Implicit Surface Reconstruction, Noisy Point Clouds}

  \vskip 0.3in
]



\printAffiliationsAndNotice{\icmlEqualContribution}  

\begin{abstract}
  Reconstructing accurate implicit surface representations from point clouds remains a challenging task, particularly when data is captured using low-quality scanning devices. These point clouds often contain substantial noise, leading to inaccurate surface reconstructions. Inspired by the Noise2Noise paradigm for 2D images, we introduce NoiseSDF2NoiseSDF, a novel method designed to extend this concept to 3D neural fields. Our approach enables learning clean neural SDFs from noisy point clouds through noisy supervision by minimizing the MSE loss between noisy SDF representations, allowing the network to implicitly denoise and refine surface estimations. We evaluate the effectiveness of NoiseSDF2NoiseSDF on benchmarks, including the ShapeNet, ABC, Famous, and Real datasets. Experimental results demonstrate that our framework significantly improves surface reconstruction quality from noisy inputs.
\end{abstract}

\section{Introduction}
\label{sec:introduction}
\begin{figure*}[!t]
  \vskip 0.2in
  \begin{center}
    \centerline{\includegraphics[width=0.8\textwidth]{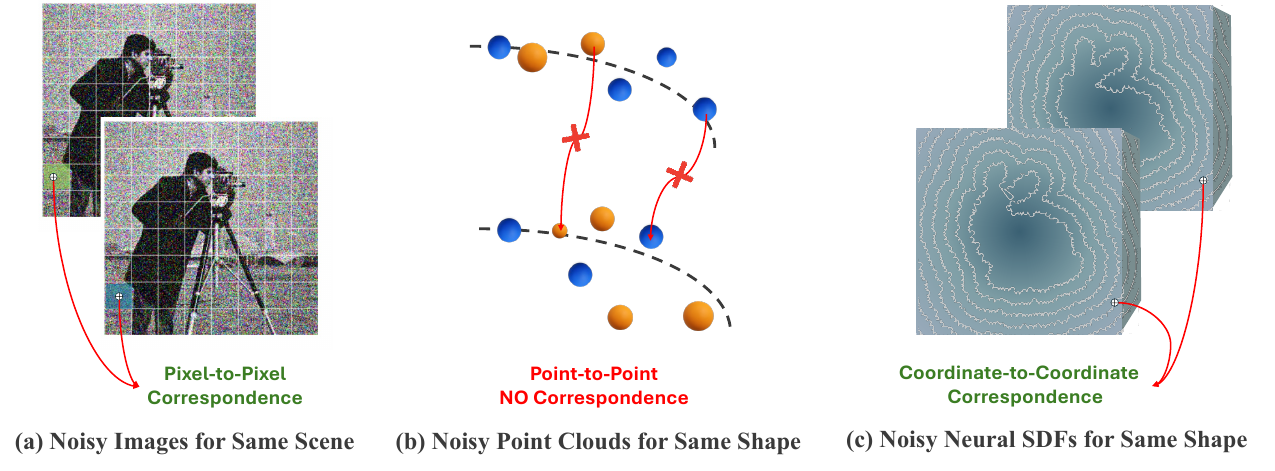}}
    \caption{
      Comparison of coordinate correspondences: (a) Pixel coordinates represent correspondences between two noisy images of the same scene. In contrast, (b) point coordinates do not exhibit correspondences between two noisy point clouds of the same shape. (c) SDF coordinates establish correspondences between two noisy neural fields representing the same shape.
    }
    \label{fig:pixel_vs_point_vs_sdf}
  \end{center}
\end{figure*}

Learning from imperfect targets \cite{zhu2017unpaired,zhang2018generalized,han2018co,pmlr-v80-lehtinen18a,hong2022pointcam,bora2018ambientgan,du2023weakly} is a fundamental challenge in machine learning, particularly when obtaining clean labels is impractical or unfeasible. In image processing, the pioneering work of Noise2Noise (N2N) \cite{pmlr-v80-lehtinen18a} demonstrated that image restoration could effectively be achieved by observing multiple corrupted instances of the same image. Specifically, N2N leverages the principle that pixel values at identical coordinates in different noisy images ideally represent the same underlying true signal. Then, the model learns to restore clean images by minimizing a simple loss, such as mean squared error (MSE), between noisy observations, as shown in Figure \ref{fig:pixel_vs_point_vs_sdf} (a).

Extending N2N principles to 3D point clouds \cite{hermosilla2019total,Ma2023SDF}, however, poses inherent limitations due to their unstructured nature. Unlike images organized on regular grids, point clouds exhibit deviations across all spatial coordinates without the benefit of a stable reference framework. This fundamental difference renders a direct extension of N2N impractical. Standard loss functions such as MSE prove ineffective, resorting to specialized loss functions like Earth Mover’s Distance (EMD) to capture only soft geometric correspondences in point cloud data, see Figure \ref{fig:pixel_vs_point_vs_sdf} (b).

Recent advances in 3D shape surface reconstruction have introduced neural fields, such as neural Signed Distance Function (SDF) \cite{park2019deepsdf,mescheder2019occupancy,cui2024numgrad,zhu2024ssp}, which are capable of predicting continuous SDF values for any given query 3D coordinate. Our key observation is that neural SDF, which encodes the 3D shape as an SDF mapping from 3D coordinates to scalar distance values, exhibits a conceptual parallel to the mapping between pixel coordinates and pixel intensities in 2D images, as shown in Figure \ref{fig:pixel_vs_point_vs_sdf} (c). Building on this analogy, we hypothesize that neural SDFs can be denoised by using noisy SDF observations with the same MSE loss strategy inspired by the N2N principle in image restoration.

In contrast to paired noisy 2D images that can be readily acquired from cameras, 3D neural fields are typically learned through neural networks. Because image coordinates are discrete and finite, simply following the image-level N2N approach and denoising the SDF values only at specified coordinates, cannot ensure that the denoised SDF field remains complete and continuous over an infinite coordinate domain. 
To address the above limitation, we propose training a neural field network with denoising capability to predict clean SDF values without being restricted to specific query coordinates. Specifically, we introduce NoiseSDF2NoiseSDF, a novel approach that employs noisy-target supervision within neural SDFs to enable robust surface reconstruction from noisy 3D point clouds. The denoising network takes independently corrupted point clouds as input and predicts the underlying clean SDF values. Instead of using clean SDFs as ground truth, we employ another noisy neural SDF, generated by off-the-shelf point-to-SDF methods, as the supervision target. We then minimize the discrepancy between the predicted SDF output and the noisy SDF target using MSE loss. Through this process, the network learns to suppress noise and improve consistency across SDF values, resulting in clean neural representations.

To evaluate the effectiveness of NoiseSDF2NoiseSDF, we conduct comprehensive experiments across benchmark datasets, including ShapeNet \cite{chang2015shapenet}, ABC \cite{koch2019abc}, Famous \cite{erler2020points2surf}, and Real \cite{erler2020points2surf}. Our experimental results demonstrate that neural SDFs can indeed be denoised effectively by employing MSE loss directly between their noisy representations. This finding confirms our central hypothesis: neural SDFs can learn to produce cleaner outputs simply by observing and minimizing discrepancies among noisy neural fields, effectively extending the N2N paradigm into the domain of 3D shape surface reconstruction. 
%
We summarize our contributions as follows:
\begin{itemize}
    \item Conceptually, our work offers a new perspective: analogous to Noise2Noise, which relies on exact pixel-wise correspondence across noisy observations, two noisy SDFs can establish exact coordinate-wise correspondence. This insight motivates our method, which reformulates the problem in the neural SDF domain.
    \item Methodologically, we address a key challenge: noisy SDF supervision is not naturally available. To overcome this, we introduce a noisy-target generation strategy based on a frozen off-the-shelf Point2SDF model, which converts noisy point clouds into noisy SDF targets and enables coordinate-wise noisy-target learning in neural SDF space.
    \item Experimentally, we demonstrate that our method remains robust across diverse noise types, target generators, and denoising backbones.
\end{itemize}

\section{Related Work}
\label{sec:related_work}

\textbf{Noise2Noise (N2N)} \cite{pmlr-v80-lehtinen18a} has significantly influenced recent image denoising. By leveraging pairs of noisy observations of the same scene,  N2N  learns to predict one noisy realization from another via pixel-wise correspondence. Subsequent methods like Noise2Void \cite{krull2019noise2void}, Noise2Self \cite{pmlr-noise2self} employ blind-spot masking techniques, training models directly on individual noisy images without pairs. Noise2Same \cite{xie2020noise2same} derives self-supervised loss bounds to eliminate the blind-spot restriction altogether. Self2Self \cite{quan2020self2self} and Neighbor2Neighbor \cite{huang2021neighbor2neighbor} exploit internal image redundancy, employing dropout or pixel resampling to train on single noisy observations without explicit noise modeling. Noisier2Noise \cite{moran2020noisier2noise} extends N2N to explicitly introduce additional synthetic noise, learning to map noisier images back to their original noisy versions.

Recent work has extended the N2N paradigm to the 3D domain, particularly to point clouds \cite{hermosilla2019total,Ma2023SDF,wang2024noise4denoise,wei2025noise2score3d}. TotalDenoising \cite{hermosilla2019total} and N2NM \cite{Ma2023SDF} employ soft local geometric correspondences with Earth Mover’s Distance (EMD) loss to align noisy point clouds with the underlying surface. However, these methods are unable to establish exact point matches. 

To the best of our knowledge, no prior work has applied the N2N paradigm to the domain of 3D neural fields. We are the first to exploit the structural similarities between neural fields and images by proposing an N2N denoising framework for 3D SDFs using a simple MSE loss, which enables direct SDF matches.

\textbf{Implicit Surface Reconstruction.}
Overfitting-based methods optimize a neural implicit function for a single shape through intensive test-time optimization. They often achieve high geometric fidelity on that specific object but lack generalization to new shapes. For example, SAL \cite{atzmon2020sal}, SALD \cite{atzmon2021sald}, and Sign-SAL \cite{zhao2021signsal} use point proximity and self-similarity cues. Gradient regularization techniques like IGR \cite{gropp2020igr}, DiGS \cite{ben2022digs}, and Neural-Pull \cite{ma2020neuralpull} improve stability and detail. Extensions such as SAP \cite{peng2021sap}, LPI \cite{chen2022lpi}, and Implicit Filtering-Net \cite{li2024implicitfiltering} enhance reconstruction under sparse sampling and complex geometry. Neural-Singular-Hessian~\cite{zixiong23neuralsingular} pushes single-shape overfitting by leveraging a Hessian-based regularizer to achieve surface recovery. While accurate, these methods are typically sensitive to noise. Robust variants (e.g., SAP \cite{peng2021sap}, PGR \cite{lin2022pgr}, Neural-IMLS \cite{wang2023neuralimls}, N2NM \cite{Ma2023SDF} and LocalN2NM \cite{chen2024localn2nm}) address this via smoothing, denoising priors, or self-supervision. 

Data-driven methods learn from collections of shapes, allowing the model to infer implicit surfaces for previously unseen instances with efficient inference. For instance, global-latent methods, such as OCCNet \cite{mescheder2019occupancy}, IM-NET \cite{chen2019imnet}, and DeepSDF \cite{park2019deepsdf}, encode entire shapes into fixed-length global latent codes. 
Local-based methods improve expressiveness by operating at finer scales. Grid-based approaches divide space into cells and learn small implicit functions per cell (ConvOccNet \cite{peng2020convocc}, SSRNet \cite{mi2020ssrnet}, Local Implicit Grid \cite{genova2020localdeep}, Deep Local Shapes \cite{chabra2020dls}). Patch-based methods segment point clouds into local regions and learn shared atomic representations (PatchNets \cite{tretschk2020patchnets}, POCO \cite{boulch2022poco}, neighborhood-based \cite{jiang2021neighborhood}). Hybrid methods combine global context with local detail. For instance, IF-Nets \cite{chibane2020ifnet} and SG-NN \cite{dai2020sgnn} integrate local features within hierarchical representations. P2S \cite{qi2017pointnet} and PPSurf \cite{erler2024ppsurf} use dual-branch networks to predict SDFs. Recent transformer-based models (ShapeFormer \cite{yan2022shapeformer}, 3DILG \cite{zhang20223dilg}, 3DS2V \cite{zhang20233dshape2vecset}, LaGeM \cite{zhang2025lagem}) leverage self-attention for long-range structure modeling. Since these methods are trained using ground-truth SDFs, their performance degrades when input point clouds are sparse or noisy. In this work, we demonstrate that clean neural fields can be learned under noisy supervision, enabling robust surface reconstruction from corrupted inputs.

\section{Preliminaries}
\label{sec:preliminaries}

In \textit{Noise2Noise} \cite{pmlr-v80-lehtinen18a}, the key idea is that given multiple noisy observations of the same underlying clean image, the pixel intensities at the same spatial coordinates are expected to share the same statistical properties. Formally, consider an image domain \( \mathbf{X} \subset \mathbb{R}^2 \), and let \( y_1, y_2, \dots, y_n \) be noisy observations of the same underlying clean image taken at different instances. For any pixel coordinate \( x \in \mathbf{X} \), the pixel intensities \( y_1(x), y_2(x), \dots, y_n(x) \) are samples drawn from a distribution centered around the true pixel value at that location, perturbed by independent, zero-mean noise. The core insight of Noise2Noise is that even in the presence of such noise, the expectation of the noisy pixel values converges to the true signal:
\begin{equation}
\mathbb{E}[y_i(x)] = y(x), \quad \forall i \in \{1, 2, \dots, n\},
\end{equation}
where \( y_i(x) \) is the observed pixel value at coordinate \( x \) in the \( i \)-th noisy image, and \( y(x) \) is the true underlying pixel value at that coordinate. This property enables training a neural network purely on noisy data, using other noisy images as supervision.

Let \( f_\theta \) denote a denoising network parameterized by \( \theta \), 
and let \( x \in \mathbf{X} \) represent a spatial query coordinate. The network is designed to predict pixel intensities given a noisy image and the query coordinate. The prediction is written as:
\begin{equation}
\hat{y}(x \mid y_i) = f_\theta(y_i, x),
\end{equation}
where \( y_i \) is the noisy input image, \( x \) is the queried pixel location, and \( \hat{y}(x \mid y_i) \) is the predicted pixel intensity at \( x \). 
The model is trained to minimize the expected squared error between the predicted pixel value and the corresponding pixel value in another independent noisy observation. The loss function is:
\begin{equation}
\mathcal{L}(\theta)
= \mathbb{E}_{\substack{y_1, y_2 \sim p(y \mid y)\\ x \sim \mathcal{U}(\mathbb{R}^2)}}
\Bigl[\, \| \hat{y}(x \mid y_1) - y_2(x) \|^2 \,\Bigr].
\end{equation}
where \( y_1, y_2 \) are independent noisy observations of the same clean image, and \( x \in \mathbf{X} \) is sampled uniformly from the image domain. After training, given a noisy image, \( f_\theta \) can predict a denoised version.

\section{Method}
\label{sec:method}

\begin{figure*}[!t]
  \vskip 0.2in
  \begin{center}
    \centerline{\includegraphics[width=0.7\linewidth]{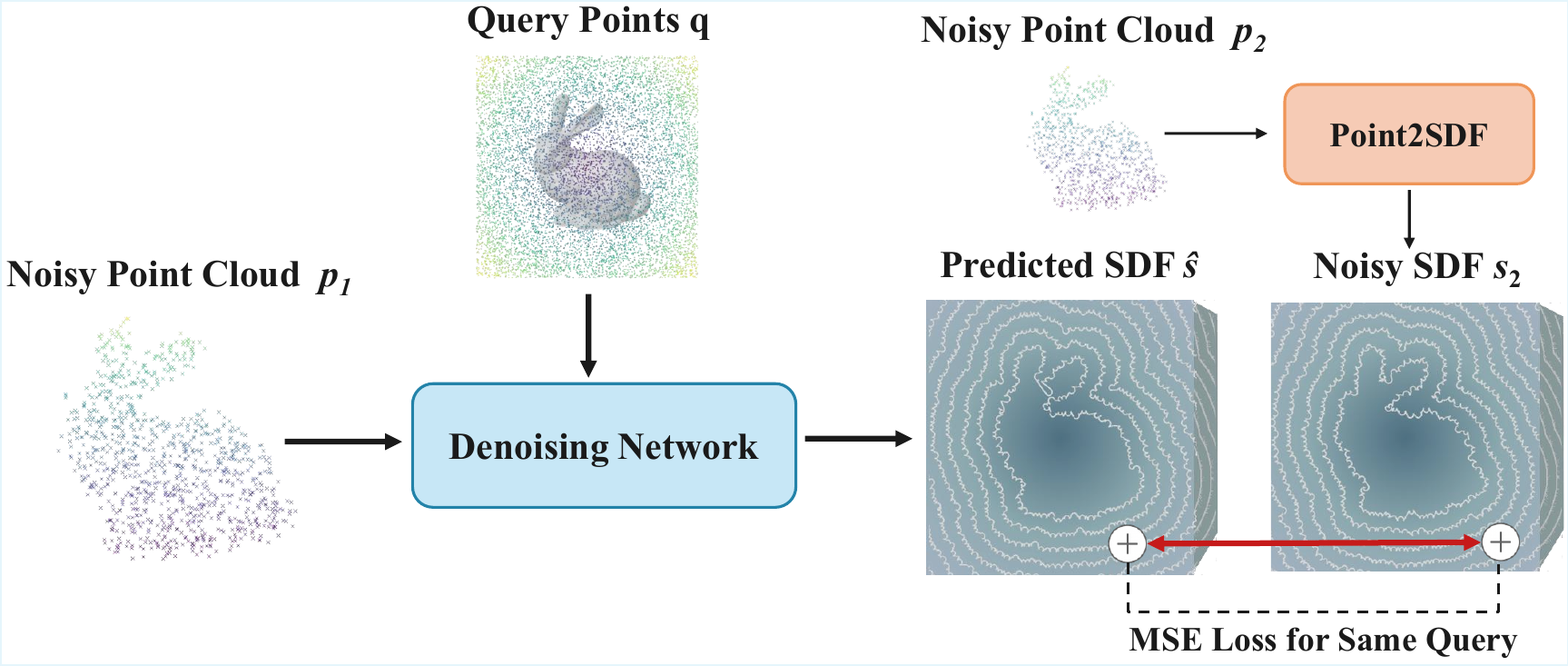}}
    \caption{The training pipeline of the NoiseSDF2NoiseSDF framework. Given two independent noisy point clouds \(p_1\) and \(p_2\) of the same underlying shape, \(p_1\) is fed into the denoising network to predict a smoothed SDF \(\hat{s}\), while \(p_2\) is passed through a Point2SDF network to generate a noisy SDF \(s_2\). Both SDFs are evaluated at a shared set of query points \(q\), and their mean squared error is used  to update the denoising network weights.}
    \label{fig:denoising_framework}
  \end{center}
\end{figure*}

Our proposed method investigates whether clean neural fields can be effectively learned by observing their noisy counterparts. Drawing inspiration from Noise2Noise, where noisy images directly serve as inputs and targets, we adapt this principle to learning neural fields from noisy point cloud data. In contrast to the direct usage of noisy images as input in traditional Noise2Noise setups, we employ a neural network conditioned on a noisy point cloud to predict neural SDFs at given query coordinates. Rather than utilizing clean SDFs as supervision, our approach leverages noisy neural fields at identical coordinates derived from another independently noisy version of the same underlying shape. This ensures one-to-one correspondence between the predicted and target neural fields, allowing effective noise suppression through direct MSE loss minimization.

\subsection{NoiseSDF2NoiseSDF}

Formulating \textit{Noise2Noise} \cite{pmlr-v80-lehtinen18a} to Signed Distance Functions (SDFs) introduces new opportunities for denoising in 3D spaces. Unlike unstructured point clouds, SDFs represent 3D geometry in a structured and continuous manner, mapping each spatial coordinate \( q \in \mathbb{R}^3 \) to its signed distance from the surface of an underlying object. This continuity ensures that, for the same query coordinate across multiple noisy observations derived from the same shape, the SDF values should remain statistically consistent. Let \( p_1, p_2, \dots, p_n \) be noisy point cloud observations of the same underlying 3D shape, and let \( s_1, s_2, \dots, s_n \) be their corresponding noisy SDFs. Given a noisy point cloud \( p_i \) and a query coordinate \( q \), a neural network \( f_\theta \), parameterized by \( \theta \), is trained to predict the SDF value \( \hat{s}(q \mid p_i) \) at the queried location:
\begin{equation}
\hat{s}(q \mid p_i) = f_\theta(p_i, q).
\end{equation}
The structured nature of SDFs enables the network to learn smooth and continuous surface representations, even from sparse or noisy inputs. This makes SDFs advantageous over unordered point clouds for tasks like 3D denoising and reconstruction.

\textbf{Training Objective.} The model is trained by minimizing the expected squared error between the predicted SDF value from one noisy observation and the SDF value at the same query location in another noisy observation of the same shape. The loss function is defined as:

\begin{equation}
\mathcal{L}(\theta)
= \mathbb{E}_{\substack{p_1,p_2 \sim p(p\mid s)\\ q \sim \mathcal{U}(\mathbb{R}^3)}}
\Bigl[\, \bigl\| \hat{s}(q \mid p_1) - s_2(q \mid p_2) \bigr\|^2 \,\Bigr].
\end{equation}

where \( p_1, p_2 \) are independent noisy point cloud observations sampled from the same underlying shape \( s \), and \( s_2(q \mid p_2) \) is the noisy SDF value at coordinate \( q \) associated with noisy point cloud \( p_2 \).

This formulation takes advantage of the continuous nature of SDFs, which, unlike point clouds, allows for consistent supervision across noisy samples even if the raw point distributions are unstructured. By learning to map noisy coordinates to structured SDF representations, the neural network effectively filters noise, yielding a refined and more accurate 3D representation of the surface.

\textbf{Analysis.} Perturbing the closest surface point with zero-mean Gaussian noise produces a noisy signed-distance estimate whose expectation remains approximately consistent with the clean signed distance.  Let $\phi$ be a signed distance function with zero level set $S$.
For a query point $q$, let $p\in S$ be its closest surface point and
let $n=\nabla \phi(p)$ be the unit normal at $p$. Since $\phi$ is an SDF,
we can locally write
\begin{equation}
q=p+s n,\qquad s=\phi(q).
\end{equation}
Suppose the closest point is perturbed by zero-mean Gaussian noise,
\begin{equation}
\tilde p=p+\varepsilon,\qquad \varepsilon\sim\mathcal N(0,\Sigma_p),
\end{equation}
and the noisy SDF target is defined by the local normal projection
\begin{equation}
\tilde s(q)=n^\top(q-\tilde p).
\end{equation}
Then $\tilde s(q)$ is an unbiased first-order estimate of the clean SDF value:
\begin{equation}
\mathbb E[\tilde s(q)] = s.
\end{equation}
In particular, the result also holds on the zero level set. The full theoretical justification is provided in Appendix.

\subsection{Implementation}
\label{sec:implementation}

Our framework is illustrated in Figure \ref{fig:denoising_framework}. The process begins with sampling sparse, noisy point clouds from a watertight surface. During training, a pair of noisy point clouds is randomly selected: one is processed through the neural SDF network to predict approximate clean SDF values for the underlying 3D shape. Simultaneously, a point-to-SDF method is applied to generate a noisy SDF target, which serves as noisy supervision during the denoising phase. For each query point, the corresponding SDF values from these two representations are extracted and compared using the Mean Squared Error (MSE) loss function. This loss is then utilized to update the weights of the neural SDF network during denoising.

\textbf{Point Sampling.} We first normalize the watertight meshes into a unit cube, then sample points from the surfaces to obtain the original point cloud \(p\). Following the Noise2Noise protocols \cite{pmlr-v80-lehtinen18a,Ma2023SDF}, we apply zero-mean Gaussian noise to generate noisy point cloud pairs. The query point set consists of \(50\%\) near-surface points and \(50\%\) uniformly sampled points from the unit cube. To reduce dependency on the original clean surface, we directly use the two input noisy point clouds as the near-surface query points. Additionally, we uniformly sample \(N\) points within the cube as spatial query points.

\textbf{Denoising Network.} Our SDF prediction network is built on 3DS2V \cite{zhang20233dshape2vecset}. Initially, a noisy point cloud \(p_{1}\) is sampled and transformed into positional embeddings, which are then encoded into a set of latent codes through a cross-attention module. Subsequently, self-attention is applied to aggregate and exchange information across the latent set, enhancing feature integration. A cross-attention module then computes interpolation weights for the query point \(q\). These interpolated feature vectors are processed through a fully connected layer to predict SDF values. The network weights are initialized following \citet{zhang20233dshape2vecset} to accelerate convergence.

\textbf{Noisy Target.} Given another paired noisy point cloud \( p_2 \), a Point2SDF method is required to predict noisy SDF values \( s_2 \) from it. In this implementation, we use the 3DS2V \cite{zhang20233dshape2vecset} with frozen parameters. Since it is not trained in a denoising setting, noisy inputs produce noisy SDF outputs. This network takes a noisy point cloud \(p_{2}\) and a query point \(q\) as input, producing the corresponding noisy SDF scalar value at \(q\). We freeze its parameters to ensure that all SDF targets are consistently noisy.

\textbf{Inference.} Only the Denoising Network is used for inference. Given a noisy point cloud sampled from an unseen 3D shape, it predicts the corresponding denoised SDF values in a single forward pass and the clean underlying surface is extracted with Marching Cubes \cite{lorensen1987marching}.

\section{Experiment}
\label{sec:experiment}

\subsection{Training Details}

We employed the AdamW optimizer~\cite{loshchilov2019decoupled} for optimization, adopting a fixed learning rate of \(1\times10^{-4}\). For resource usage, we trained on three Nvidia A100 GPUs with a batch size of 32 per GPU, taking approximately 15 hours for the ShapeNet dataset and 2.5 hours for the ABC dataset.
We sampled 2048 points from watertight meshes as the initial point cloud. Following the N2NM \cite{Ma2023SDF}, we applied Gaussian noise with standard deviations of \(1\%\) and \(2\%\) online to generate noisy and sparse point cloud pairs. Additionally, we sampled 8192 query points online. The noise magnitude is defined with respect to both the point‑cloud bounding‑box size and the point density. For a fixed numeric noise level, a smaller bounding box amplifies the relative impact of the perturbation. All point clouds are normalized to the cubes \([-0.5,0.5]^3\) or \([-1,1]^3\). Furthermore, sparser point sets are more susceptible to noise. With only 2048 points, noise levels of \(0.01\) and \(0.02\) constitute \emph{severe} perturbations irrespective of the bounding‑box scale.

\subsection{Datasets and Metrics}

We evaluated our NoiseSDF2NoiseSDF on ShapeNet following \citet{zhang20233dshape2vecset}. To assess denoising effectiveness and surface reconstruction quality, we used metrics, including Intersection-over-Union (IoU), Chamfer Distance, F1 Score, and Normal Consistency (NC). IoU was computed based on occupancy predictions over densely sampled volumetric points. Following methods \cite{ma2020neuralpull,li2024implicitfiltering}, we sampled \(1 \times 10^5\) points from the reconstructed and ground-truth surfaces to compute the Chamfer Distance and F1 Score.

To further evaluate the generalization ability of our approach, we trained the model on the ABC training set \cite{koch2019abc} and tested it on the ABC test set, as well as the Famous \cite{erler2020points2surf} and Real \cite{erler2020points2surf} datasets. Importantly, neither of these datasets was used during 3DS2V’s training, so they can be regarded as out-of-distribution. Our model is trained solely with noisy targets generated by 3DS2V, without relying on any clean ground-truth from these datasets. We utilized the preprocessed datasets and data splits provided by \citet{erler2020points2surf,erler2024ppsurf}. We reported evaluation metrics including Normal Consistency, Mesh Normal Consistency, Chamfer Distance, and F1 Score. All metrics reported above are evaluated on the reconstructed meshes. We excluded IoU from this benchmark because, under severe noise, many reconstructed meshes become non-watertight or heavily degenerated, making it infeasible to assign reliable inside/outside labels and rendering the IoU metric unreliable.

\subsection{Results on ShapeNet}

\begin{table*}[!t]
\captionsetup{ width=\textwidth} 
\caption{Comparison of 3DS2V \cite{zhang20233dshape2vecset} and Ours on ShapeNet with Gaussian noise $\sigma = 0.01$. 
}
\label{tab:shapenet_sigma001_comparison}
\centering
\footnotesize
\setlength{\tabcolsep}{4pt}
\renewcommand{\arraystretch}{1.1}
\rowcolors{2}{blue!5}{white}
\begin{adjustbox}{width=0.9\textwidth}
\begin{tabular}{lcccccccccccc}
\toprule
\textbf{Category} &
\multicolumn{3}{c}{\textbf{IoU} \(\uparrow\)} &
\multicolumn{3}{c}{\textbf{NC} \(\uparrow\)} &
\multicolumn{3}{c}{\textbf{Chamfer} \(\downarrow\)} &
\multicolumn{3}{c}{\textbf{F-Score} \(\uparrow\)} \\ 
\cmidrule(lr){2-4}\cmidrule(lr){5-7}\cmidrule(lr){8-10}\cmidrule(lr){11-13}
& 3DS2V & Ours & $\Delta$
& 3DS2V & Ours & $\Delta$
& 3DS2V & Ours & $\Delta$
& 3DS2V & Ours & $\Delta$ \\
\midrule
table      & 0.879 & \textbf{0.922} & +0.043 & 0.930 & \textbf{0.976} & +0.046 & 0.013 & \textbf{0.012} & +0.001 & 0.991 & \textbf{0.992} & +0.001 \\
car        & 0.946 & \textbf{0.959} & +0.013 & 0.890 & \textbf{0.908} & +0.018 & 0.022 & \textbf{0.020} & +0.002 & 0.925 & \textbf{0.925} & +0.000 \\
chair      & 0.887 & \textbf{0.921} & +0.034 & 0.937 & \textbf{0.966} & +0.029 & 0.014 & \textbf{0.013} & +0.001 & 0.986 & \textbf{0.986} & +0.000 \\
airplane   & 0.884 & \textbf{0.931} & +0.047 & 0.939 & \textbf{0.972} & +0.033 & 0.010 & \textbf{0.008} & +0.002 & 0.997 & \textbf{0.997} & +0.000 \\
sofa       & 0.946 & \textbf{0.964} & +0.018 & 0.943 & \textbf{0.974} & +0.031 & 0.014 & \textbf{0.012} & +0.002 & 0.986 & \textbf{0.987} & +0.001 \\
rifle      & 0.821 & \textbf{0.910} & +0.089 & 0.869 & \textbf{0.960} & +0.091 & 0.009 & \textbf{0.007} & +0.002 & 0.997 & \textbf{0.998} & +0.001 \\
lamp       & 0.826 & \textbf{0.894} & +0.068 & 0.904 & \textbf{0.952} & +0.048 & 0.011 & \textbf{0.009} & +0.002 & 0.989 & \textbf{0.989} & +0.000 \\
\midrule
mean & 0.884 & \textbf{0.929} & +0.045 & 0.916 & \textbf{0.958} & +0.042 & 0.0132 & \textbf{0.0113} & +0.0019 & 0.981 & \textbf{0.986} & +0.005 \\
\bottomrule
\end{tabular}
\end{adjustbox}
\end{table*}

\begin{table*}[!t]
\captionsetup{ width=\textwidth}
\caption{Comparison of 3DS2V \cite{zhang20233dshape2vecset} and Ours on ShapeNet with Gaussian noise $\sigma = 0.02$.
}
\label{tab:shapenet_sigma002_comparison}
\centering

    \footnotesize
\setlength{\tabcolsep}{4pt}
\renewcommand{\arraystretch}{1.1}
\rowcolors{2}{blue!5}{white}

\begin{adjustbox}{width=0.9\textwidth}
\begin{tabular}{lcccccccccccc}
\toprule
\textbf{Category} &
\multicolumn{3}{c}{\textbf{IoU} \(\uparrow\)} &
\multicolumn{3}{c}{\textbf{NC} \(\uparrow\)} &
\multicolumn{3}{c}{\textbf{Chamfer} \(\downarrow\)} &
\multicolumn{3}{c}{\textbf{F-Score} \(\uparrow\)} \\ 
\cmidrule(lr){2-4}\cmidrule(lr){5-7}\cmidrule(lr){8-10}\cmidrule(lr){11-13}
& 3DS2V & Ours & $\Delta$
& 3DS2V & Ours & $\Delta$
& 3DS2V & Ours & $\Delta$
& 3DS2V & Ours & $\Delta$ \\
\midrule
table      & 0.528 & \textbf{0.591} & +0.063 & 0.765 & \textbf{0.912} & +0.147 & 0.029 & \textbf{0.028} & +0.001 & 0.792 & \textbf{0.859} & +0.067 \\
car        & 0.434 & \textbf{0.491} & +0.057 & 0.715 & \textbf{0.787} & +0.072 & \textbf{0.040} & 0.044 & -0.004 & 0.669 & \textbf{0.688} & +0.019 \\
chair      & 0.463 & \textbf{0.530} & +0.067 & 0.729 & \textbf{0.868} & +0.139 & \textbf{0.034}& 0.035 & -0.001 & 0.694 & \textbf{0.721} & +0.027 \\
airplane   & 0.465 & \textbf{0.536} & +0.071 & 0.719 & \textbf{0.856} & +0.137 & 0.025 & \textbf{0.022} & +0.003 & 0.830 & \textbf{0.899} & +0.069 \\
sofa       & 0.355 & \textbf{0.425} & +0.070 & 0.769 & \textbf{0.866} & +0.097 & \textbf{0.036} & 0.038 & -0.002 & 0.667 & \textbf{0.677} & +0.010 \\
rifle      & 0.625 & \textbf{0.781} & +0.156 & 0.691 & \textbf{0.891} & +0.200 & 0.021 & \textbf{0.014} & +0.007 & 0.887 & \textbf{0.968} & +0.081 \\
lamp       & 0.572 & \textbf{0.649} & +0.077 & 0.744 & \textbf{0.896} & +0.152 & 0.026 & \textbf{0.025} & +0.001 & 0.825 & \textbf{0.880} & +0.055 \\
\midrule
mean & 0.492 & \textbf{0.572} & +0.080 & 0.733 & \textbf{0.868} & +0.135 & 0.030 & \textbf{0.029} & +0.001 & 0.766 & \textbf{0.813} & +0.047 \\
\bottomrule
\end{tabular}
\end{adjustbox}
\end{table*}

To verify our hypothesis that clean neural fields can be learned from noisy supervision, we comprehensively compared our method with 3DS2V \cite{zhang20233dshape2vecset} on the seven largest ShapeNet \cite{chang2015shapenet} subsets, following its experimental setup. Since our method is trained using 3DS2V as the denoising network with only noisy supervision, achieving superior performance over 3DS2V would therefore provide strong evidence supporting our hypothesis.

We adopt the officially released 3DS2V model as our baseline. 3DS2V was trained with pairs of clean point clouds and clean SDFs for supervision, and it was not exposed to noisy inputs paired with clean SDF. To ensure a fair evaluation, we adopt the same data splits and preprocessing procedures as 3DS2V. Our NoiseSDF2NoiseSDF is trained under noisy supervision, meaning that we do not use any paired noisy point clouds and clean SDFs throughout the entire training process.

\begin{figure*}[!t]
  \vskip 0.2in
  \begin{center}
    \centerline{\includegraphics[width=1\textwidth]{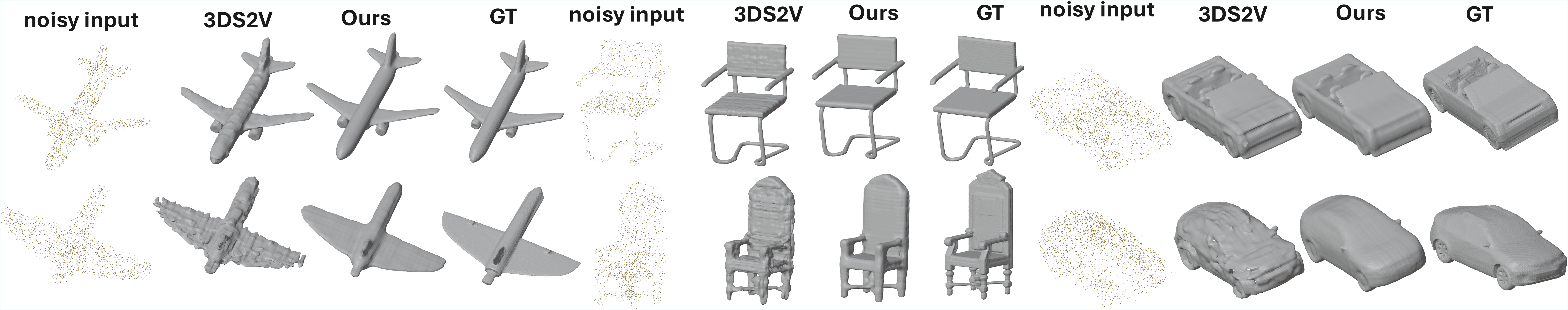}}
    \caption{Comparison on the ShapeNet dataset. The first row shows Gaussian noise with \(\sigma = 0.01\), and the second row with \(\sigma = 0.02\). Ours produces smoother reconstructions, better aligning with the underlying surfaces compared to baseline.
    }
    \label{fig:shapenet_reconstruction_compare}
  \end{center}
\end{figure*}

We reported evaluation results for each subset at noise levels of 0.01 (Table \ref{tab:shapenet_sigma001_comparison}) and 0.02 (Table \ref{tab:shapenet_sigma002_comparison}) and showed visualization results in Figure \ref{fig:shapenet_reconstruction_compare}. Under lower corruption (\(\sigma = 0.01\)), our method outperforms 3DS2V across all evaluation metrics. For example, the mean results show an IoU increase of 0.045 (5.1\%), a Normal Consistency improvement of 0.042 (4.6\%), and a reduction in Chamfer Distance from 0.0132 to 0.0113. At the higher corruption level (\(\sigma = 0.02\)), our approach remains the robust and stable with the better mean metrics that surpass the baseline. For instance, the mean results show an IoU increase of 0.08 (16.3\%), a Normal Consistency improvement of 0.135 (18.4\%), and an F-Score increase of 0.047 (6\%). The results demonstrate that our method outperforms the baseline, 3DS2V, under both corruption levels and confirm our central idea that it is possible to learn to clean neural fields from noisy supervision.

\subsection{Results on ABC, Famous, and Real}

\begin{figure*}[!t]
  \vskip 0.2in
  \begin{center}
    \centerline{\includegraphics[width=\textwidth]{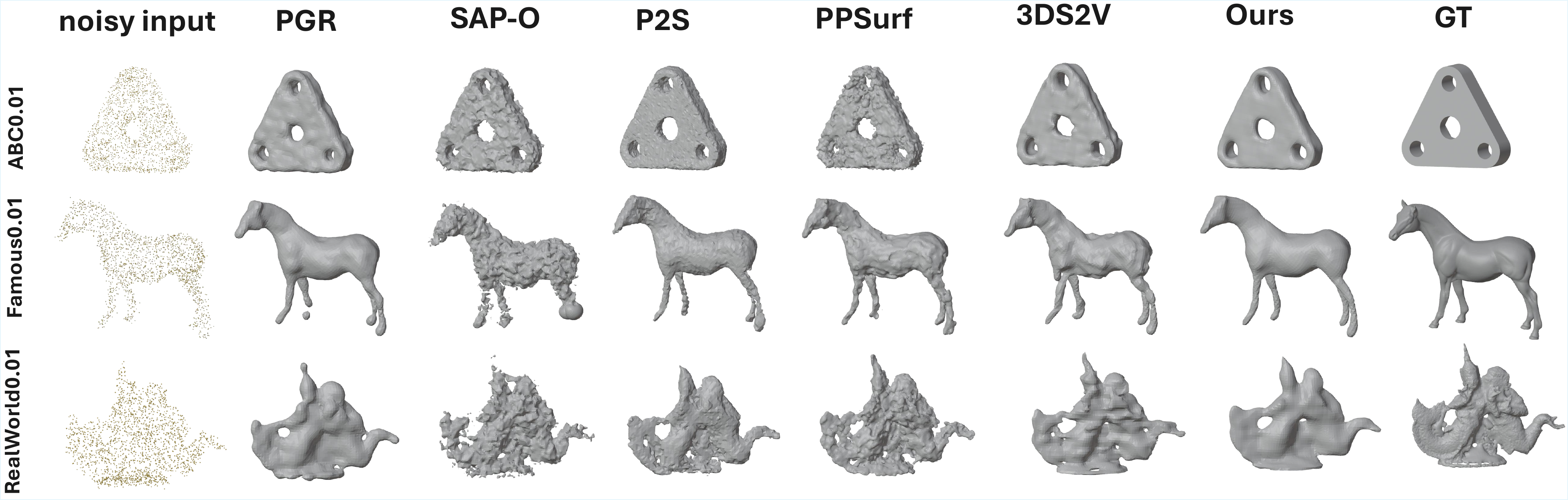}}
    \caption{Visualization of surface reconstruction under Gaussian noise (\(\sigma = 0.01\) and \(0.02\)), comparing two overfitting-based methods (PGR \cite{lin2022pgr} and SAP-O \cite{peng2021sap}), three data-driven approaches (P2S \cite{erler2020points2surf}, PPSurf \cite{erler2024ppsurf}, 3DS2V \cite{zhang20233dshape2vecset}), and our proposed method. Results are shown for three benchmark datasets: ABC \cite{koch2019abc}, Famous \cite{erler2020points2surf}, and Real \cite{erler2020points2surf}.
    }
    \label{fig:abc_reconstruction_compare}
  \end{center}
\end{figure*}

We compared results on the ABC, Famous, and Real test datasets provided by P2S \cite{erler2020points2surf}. Specifically, we evaluated data-driven methods P2S, PPSurf \cite{erler2024ppsurf}, POCO \cite{boulch2022poco}, and NKSR \cite{huang2023nksr}, known for their strong noise resilience in point cloud data. Note that, except for our method, \textit{all other approaches are trained on the ABC dataset directly using ground-truth noisy–clean pairs}. For these methods, we used their officially released pretrained models. Quantitative results are reported in Table \ref{tab:abc_datadriven_compare}, and qualitative mesh reconstructions are shown in Figure \ref{fig:abc_reconstruction_compare}. 
%
%
Across noise levels, our method achieves strong performance on mean NC and Mesh NC, indicating coherent geometry and smooth surfaces; this is also evident in the visual reconstructions (Figure \ref{fig:abc_reconstruction_compare}). At the 0.01 noise level, the ABC and Famous datasets achieve NC scores of 0.865 and 0.831, respectively. When the noise level increases to 0.02, our method yields the best mean Mesh NC and strong mean NC among them. These results demonstrate that even when trained with noisy supervision and without ground-truth, our method can achieve performance competitive with SOTA data-driven approaches.

\subsection{Quantitative Results Supplementary}

\begin{table*}[!t]
\captionsetup{justification=raggedright, width=\textwidth, skip=10pt}
\caption{Comparison of P2S \cite{erler2020points2surf}, PPSurf \cite{erler2024ppsurf}, POCO \cite{boulch2022poco}, NKSR \cite{huang2023nksr}, and Ours on six noisy test datasets. The best two performances are highlighted.}
\label{tab:abc_datadriven_compare}
\centering
\footnotesize
\setlength{\tabcolsep}{1.5pt}
\renewcommand{\arraystretch}{1.3}
\rowcolors{2}{blue!5}{white}
\begin{adjustbox}{max width=\textwidth}
\begin{tabular}{lcccccccccccccccccccccccc}
\toprule
\textbf{Dataset} & 
\multicolumn{5}{c}{\textbf{NC} \(\uparrow\)} & 
\multicolumn{5}{c}{\textbf{Mesh NC} \(\downarrow\)} & 
\multicolumn{5}{c}{\textbf{Chamfer} \(\downarrow\)} & 
\multicolumn{5}{c}{\textbf{F-Score} \(\uparrow\)} \\ 
\cmidrule(lr){2-6}\cmidrule(lr){7-11}\cmidrule(lr){12-16}\cmidrule(lr){17-21}
& P2S & PPSurf & POCO & NKSR & Ours
& P2S & PPSurf & POCO & NKSR & Ours
& P2S & PPSurf & POCO & NKSR & Ours
& P2S & PPSurf & POCO & NKSR & Ours\\
\midrule
ABC \((\sigma=0.01)\)& 0.790 & 0.770 & \underline{0.864} & 0.800 & \textbf{0.865}
                     & 0.330 & 0.059 & \underline{0.025} & 0.039 & \textbf{0.024}
                     & 0.017 & 0.017 & \textbf{0.014} & 0.018 & \underline{0.015}
                     & 0.919 & 0.935 & \textbf{0.941} & 0.927 & \underline{0.938} \\
ABC \((\sigma=0.02)\)& 0.753 & 0.728 & \textbf{0.848} & 0.727 & \underline{0.812}
                     & 0.381 & 0.061 & \underline{0.020} & 0.058 & \textbf{0.018}
                     & 0.027 & \textbf{0.022} & \underline{0.019} & 0.026 & 0.032
                     & 0.852 & 0.870 & \textbf{0.898} & 0.801 & \underline{0.724} \\
Famous \((\sigma=0.01)\)& 0.771 & 0.761 & \underline{0.825} & 0.775 & \textbf{0.831}
                     & 0.268 & 0.053 & \underline{0.026} & 0.032 & \textbf{0.025}
                     & 0.017 & \textbf{0.015} & 0.017 & 0.017 & \underline{0.016}
                     & 0.928 & \underline{0.959} & \textbf{0.962} & 0.943 & 0.941 \\
Famous \((\sigma=0.02)\)& 0.727 & 0.728 & \textbf{0.785} & 0.703 & \underline{0.767}
                     & 0.328 & 0.054 & \textbf{0.023} & 0.054 & \underline{0.024}
                     & 0.022 & \textbf{0.020} & \underline{0.022} & 0.026 & 0.032
                     & 0.868 & \textbf{0.899} & \underline{0.871} & 0.810 & 0.726 \\
Real \((\sigma=0.01)\)& 0.789 & 0.776 & \underline{0.845} & 0.779 & \textbf{0.845}
                     & 0.177 & 0.057 & \underline{0.032} & 0.038 & \textbf{0.031}
                     & 0.016 & 0.016 & \textbf{0.014} & 0.015 & \underline{0.015}
                     & 0.946 & 0.954 & \textbf{0.964} & 0.955 & \underline{0.956} \\
Real \((\sigma=0.02)\)& 0.734 & 0.745 & \textbf{0.803} & 0.700 & \underline{0.793}
                     & 0.269 & 0.053 & \underline{0.024} & 0.059 & \textbf{0.020}
                     & \textbf{0.021} & \underline{0.022} & 0.025 & 0.029 & 0.026
                     & \underline{0.877} & 0.876 & \textbf{0.930} & 0.822 & 0.809 \\
\midrule
mean (\(\sigma=0.01\)) & 0.783 & 0.769 & \underline{0.844} & 0.785 & \textbf{0.847}
                     & 0.258 & 0.056 & \underline{0.028} & 0.036 & \textbf{0.027}
                     & 0.017 & \textbf{0.014} & 0.016 & 0.017 & \underline{0.015}
                     & 0.931 & \underline{0.949} & \textbf{0.962} & 0.930 & 0.945 \\
mean (all)           & 0.761 & 0.751 & \textbf{0.828} & 0.747 & \underline{0.819}
                     & 0.292 & 0.056 & \underline{0.025} & 0.047 & \textbf{0.024}
                     & 0.020 & \textbf{0.019} & \underline{0.019} & 0.022 & 0.023
                     & 0.898 & \textbf{0.916} & \underline{0.915} & 0.876 & 0.849 \\
\bottomrule
\end{tabular}
\end{adjustbox}
\end{table*}

We compared our method with representative overfitting-based approaches, such as SAP-O \cite{peng2021sap} and PGR \cite{lin2022pgr}. These methods train separate networks for each test shape, require long inference times, and lack generalization to unseen shapes. We adopted the training configurations recommended or set as default in their respective works. Table \ref{tab:abc_overfitting_compare} shows that SAP-O and PGR can sometimes achieve lower Chamfer Distance and higher F1, but our approach consistently outperforms them in NC and Mesh NC. At the 0.01 noise level, our method achieves average NC/Mesh NC/F-Score values of 0.847/0.027/0.945, all of which are the best scores. At the 0.02 noise level, our method also demonstrates strong performance, with higher NC and Mesh NC, as well as competitive Chamfer Distance and F-Score.

\begin{table*}[!t]
   \captionsetup{justification=raggedright, width=\textwidth, skip=10pt}
    \caption{Comparison of SAP‑O \cite{peng2021sap}, PGR \cite{lin2022pgr}, and Ours on six noisy test datasets. The adaptive resolution in PGR can occasionally drop relatively low, resulting in artificially smooth meshes.}
  \label{tab:abc_overfitting_compare}
  \centering

    \footnotesize
  \setlength{\tabcolsep}{4pt}
  \renewcommand{\arraystretch}{1.1}
  \rowcolors{2}{blue!5}{white}
  \begin{adjustbox}{max width=0.9\textwidth}
  \begin{tabular}{lcccccccccccc}
    \toprule
    \textbf{Dataset} &
    \multicolumn{3}{c}{\textbf{NC} \(\uparrow\)} &
    \multicolumn{3}{c}{\textbf{Mesh NC} \(\downarrow\)} &
    \multicolumn{3}{c}{\textbf{Chamfer} \(\downarrow\)} &
    \multicolumn{3}{c}{\textbf{F‑Score} \(\uparrow\)}  \\
    \cmidrule(lr){2-4}\cmidrule(lr){5-7}\cmidrule(lr){8-10}\cmidrule(lr){11-13}
     & SAP‑O & PGR & Ours
     & SAP‑O & PGR & Ours
     & SAP‑O & PGR & Ours
     & SAP‑O & PGR & Ours \\
    \midrule
    ABC \((\sigma=0.01)\)   & 0.710 & 0.835 & \textbf{0.865}
                    & 0.079 & 0.037 & \textbf{0.024}
                    & 0.021 & 0.020 & \textbf{0.014}
                    & 0.906 & 0.896 & \textbf{0.938} \\
    ABC \((\sigma=0.02)\)   & 0.622 & 0.778 & \textbf{0.812}
                    & 0.095 & 0.065 & \textbf{0.018}
                    & 0.026 & \textbf{0.026} & 0.032
                    & \textbf{0.824} & 0.815 & 0.724 \\
    Famous \((\sigma=0.01)\)& 0.745 & 0.813 & \textbf{0.831}
                    & 0.053 & 0.035 & \textbf{0.025}
                    & 0.022 & 0.017 & \textbf{0.016}
                    & 0.876 & 0.931 & \textbf{0.941} \\
    Famous \((\sigma=0.02)\)& 0.614 & 0.755 & \textbf{0.767}
                    & 0.104 & 0.064 & \textbf{0.024}
                    & \textbf{0.023} & 0.024 & 0.032
                    & 0.849 & 0.834 & \textbf{0.726} \\
    Real \((\sigma=0.01)\)  & 0.683 & 0.827 & \textbf{0.845}
                    & 0.097 & 0.032 & \textbf{0.031}
                    & 0.023 & 0.015 & \textbf{0.015}
                    & 0.902 & 0.956 & \textbf{0.956} \\
    Real \((\sigma=0.02)\)  & 0.595 & 0.756 & \textbf{0.793}
                    & 0.122 & 0.062 & \textbf{0.020}
                    & \textbf{0.025} & 0.026 & 0.026
                    & \textbf{0.841} & 0.824 & 0.809 \\
    \midrule
    mean (\(\sigma=0.01\))     & 0.713 & 0.825 & \textbf{0.847}  
                    & 0.076 & 0.035 & \textbf{0.027} 
                    & 0.022 & 0.017 & \textbf{0.015}  
                    & 0.895 & 0.928 & \textbf{0.945} \\
    mean (all)      & 0.661 & 0.794 & \textbf{0.819}  
                    & 0.092 & 0.049 & \textbf{0.024}  
                    & 0.023 & \textbf{0.021} & 0.023  
                    & 0.866 & \textbf{0.876} & 0.849 \\
    \bottomrule
  \end{tabular}
  \end{adjustbox}
\end{table*}

\begin{table}[htbp]
\centering
\captionsetup{justification=raggedright, width=\linewidth, skip=10pt}

\caption{Comparison of N2NM and our method under medium noise (MedN) and maximum noise (MaxN) using Chamfer Distance on the Famous dataset.}
\scriptsize
\rowcolors{2}{blue!5}{white}
\begin{tabular}{lccr}
\toprule
\textbf{Method} & MedN Chamfer $\downarrow$ & MaxN Chamfer $\downarrow$ & Inference time \\
\midrule
N2NM      & 0.0132 & \textbf{0.0231} & 2760 seconds \\
Ours      & 0.0160 & 0.0320  & \textbf{0.05 seconds}\\
Ours-TTO  & \textbf{0.0108} & 0.0252 &  20 seconds\\
\bottomrule
\end{tabular}
\label{tab:compare_n2nm_tto}
\end{table}

We further included a comparison with N2NM \cite{Ma2023SDF}. For each shape, N2NM was trained with 200 noisy samples, which resulted in long inference times (46 minutes) per shape. To ensure a fair comparison, we conducted experiments on the Famous dataset following the N2NM setup \cite{zhou2024fast} and incorporated test-time optimization (TTO) into our method with the same number of noisy samples. As shown in Table \ref{tab:compare_n2nm_tto}, our method combined with TTO demonstrated significant self-improvement, outperformed N2NM under medium noise conditions, and achieved comparable results under maximum noise, while delivering inference that is orders of magnitude faster.

\begin{table*}[htbp]
     \captionsetup{justification=centering, width=\textwidth, skip=10pt}
    \caption{Comparison between the 3DS2V \cite{zhang20233dshape2vecset} and Ours under various noise types.}
    \label{tab:noisetype}
    \centering
    \footnotesize
    \setlength{\tabcolsep}{4pt}
    \renewcommand{\arraystretch}{1.1}
    \rowcolors{2}{blue!5}{white}
    \begin{adjustbox}{max width=0.92\textwidth}
    \begin{tabular}{lcccccccccccc}
        \toprule
        \textbf{Dataset} &
        \multicolumn{3}{c}{\textbf{IoU} \(\uparrow\)} &
        \multicolumn{3}{c}{\textbf{NC} \(\uparrow\)} &
        \multicolumn{3}{c}{\textbf{Chamfer} \(\downarrow\)} &
        \multicolumn{3}{c}{\textbf{F-Score} \(\uparrow\)} \\
        \cmidrule(lr){2-4} \cmidrule(lr){5-7} \cmidrule(lr){8-10} \cmidrule(lr){11-13}
         & 3DS2V & Ours & $\Delta$
         & 3DS2V & Ours & $\Delta$
         & 3DS2V & Ours & $\Delta$
         & 3DS2V & Ours & $\Delta$ \\
        \midrule
        Uniform \((\sigma=0.01)\)  & 0.873 & \textbf{0.911} & +0.038 & 0.920 & \textbf{0.962} & +0.042 & 0.015 & \textbf{0.013} & +0.002 & 0.985 & \textbf{0.986} & +0.001 \\
        Discrete \((\sigma=0.01)\) & 0.867 & \textbf{0.895} & +0.028 & 0.915 & \textbf{0.960} & +0.045 & 0.015 & \textbf{0.014} & +0.001 & 0.984 & \textbf{0.985} & +0.001 \\
        Laplace \((\sigma=0.01)\)  & \textbf{0.909} & 0.908 & -0.001 & 0.956 & \textbf{0.956} & +0.000 & 0.014 & \textbf{0.014} & +0.000 & 0.985 & \textbf{0.985} & +0.000 \\
        Gaussian \((\sigma=0.01)\) & 0.887 & \textbf{0.927} & +0.040 & 0.937 & \textbf{0.966} & +0.029 & 0.014 & \textbf{0.013} & +0.001 & 0.986 & \textbf{0.986} & +0.000 \\
        Gaussian \((\sigma=0.01, \mu=0.005)\) & 0.860 & \textbf{0.881} & +0.021 & 0.942 & \textbf{0.964} & +0.022 & 0.017 & \textbf{0.016} & +0.001 & 0.982 & \textbf{0.984} & +0.002 \\
        Gaussian \((\sigma=0.01, \mu=0.01)\)  & 0.798 & \textbf{0.800} & +0.002 & 0.946 & \textbf{0.949} & +0.003 & 0.024 & \textbf{0.023} & +0.001 & 0.952 & \textbf{0.960} & +0.008 \\
        Gaussian \((\sigma=0.01, \mu=0.02)\)  & 0.661 & \textbf{0.666} & +0.005 & 0.875 & \textbf{0.898} & +0.023 & \textbf{0.038} & 0.039 & -0.001 & \textbf{0.549} & 0.524 & -0.025 \\
        \midrule
        mean & 0.836 & \textbf{0.855} & +0.019 & 0.927 & \textbf{0.950} & +0.023 & 0.020 & \textbf{0.019} & +0.001 & 0.917 & \textbf{0.912} & -0.005 \\
        \midrule
    \end{tabular}
    \end{adjustbox}
\end{table*}

\subsection{Ablation Study}
In the ablation setting, we evaluated our design choices on the “Chair” subset of ShapeNet, with 6271 models for training, 169 for validation, and 338 for testing.

\textbf{Noise Type.} Beyond standard zero‐mean Gaussian noise, we evaluated three additional noise types, Uniform, Discrete, and Laplace noise, each applied at a fixed magnitude of \(\sigma=0.01\). Furthermore, to assess the impact of non-zero bias in Gaussian perturbations, we conducted experiments over the domain \([-1,1]^3\) using means of \(\mu = 0.005\), \(0.01\), and \(0.02\). Comprehensive quantitative results are presented in Table \ref{tab:noisetype}. At \(\sigma=0.01\) on \([-1,1]^3\), our model consistently shows denoising performance under Uniform and Discrete noise, with notable gains in both IoU and NC.

\textbf{Noisy vs. Clean Supervision.} Under the Noise2Noise (N2N) setting, clean supervision reduces to Noise2Clean, where ground-truth SDF values are provided during training. For a fair comparison, we conducted experiments under identical training conditions (batch size, shapes numbers/batch, etc.) between our noisy and clean supervision with a noise level of 0.01. The results are reported in Table \ref{tab:supervision_signal}~(Clean as target). We observed that our noisy supervision achieves performance nearly equivalent to that of clean supervision, which aligns with the findings of \citet{pmlr-v80-lehtinen18a}. This supports our main hypothesis that cleaner outputs can be produced by observing noisy neural fields.

\textbf{Noisy Targets and Denoising Networks.} To validate that our framework can relax constraints on the architecture of the SDF prediction (i.e., Point2SDF), we conducted a study by replacing 3DS2V with 3DILG as the noisy targets. We further used 3DILG as the denoising network and performed the same experiments. 3DILG \cite{zhang20223dilg} encodes 3D shapes using irregular latent grids, whereas 3DS2V represents neural fields with a set of vectors. In Table \ref{tab:supervision_signal}, the results indicate that our method remains effective in learning clean SDFs from noisy supervision. These findings support the generalizability of our approach across different forms of noisy supervision and denoising networks.

\begin{table}[t]
\centering
\small
\captionsetup{width=\linewidth, skip=10pt}
\caption{Quantitative comparisons for different targets and denoising networks.}
\label{tab:supervision_signal}

\begin{subtable}[t]{\linewidth}
\centering
\caption{Using 3DS2V as denoising network.}
\begin{tabular}{lcccc}
\toprule
\textbf{Method} & \textbf{IoU $\uparrow$} & \textbf{Chamfer $\downarrow$} & \textbf{F-Score $\uparrow$} & \textbf{NC $\uparrow$} \\
\midrule
Baseline     & 0.887 & 0.014 & 0.986 & 0.937 \\
\midrule
Clean        & 0.939 & 0.013 & 0.988 & 0.970 \\
Noisy(3DILG) & 0.903 & 0.014 & 0.980 & 0.950 \\
Noisy(3DS2V) & 0.927 & 0.013 & 0.986 & 0.966 \\
\bottomrule
\end{tabular}
\end{subtable}

\vspace{6pt}

\begin{subtable}[t]{\linewidth}
\centering
\caption{Using 3DILG as denoising network.}
\begin{tabular}{lcccc}
\toprule
\textbf{Method} & \textbf{IoU $\uparrow$} & \textbf{Chamfer $\downarrow$} & \textbf{F-Score $\uparrow$} & \textbf{NC $\uparrow$} \\
\midrule
Baseline & 0.881  & 0.015  & 0.977  & 0.930  \\
\midrule
Clean  & 0.921 &  0.014 & 0.981 & 0.960     \\
Noisy(3DILG) & 0.907 & 0.014 & 0.979 & 0.953 \\
Noisy(3DS2V) & 0.913 & 0.014 & 0.978 & 0.962 \\
\bottomrule
\end{tabular}
\end{subtable}

\end{table}

\textbf{Denoising Network Training.} 

Fine-tuning only the fully connected layer offers virtually no benefit. Adding the cross-attention block introduces a clear gain. Fine-tuning the entire decoder achieves better performance. We further evaluated the random initialization of the decoder and the entire network. Table \ref{tab:fine-tuning} shows that initializing the decoder from 3DS2V does not improve performance but does accelerate convergence. Considering the trade-off between performance and training cost, we adopt the strategy of freezing the encoder while fine-tuning the decoder.

\begin{table}[t]
  \captionsetup{width=\linewidth, skip=10pt}
  \caption{Comparison of training strategies on the denoising network.}
  \label{tab:fine-tuning}
  \centering
  \footnotesize
  \setlength{\tabcolsep}{2pt}
  \renewcommand{\arraystretch}{1.1}

  \begin{tabular*}{\linewidth}{@{}@{\extracolsep{\fill}}
    >{\centering\arraybackslash}m{0.11\linewidth}
    >{\centering\arraybackslash}m{0.12\linewidth}
    >{\centering\arraybackslash}m{0.07\linewidth}
    >{\centering\arraybackslash}m{0.11\linewidth}
    >{\centering\arraybackslash}m{0.11\linewidth}
    >{\centering\arraybackslash}m{0.17\linewidth}
    >{\centering\arraybackslash}m{0.17\linewidth}
  @{}}
    \toprule
    \textbf{Metric} &
    \textbf{Baseline} &
    \textbf{FC} &
    \textbf{FC+CA} &
    \textbf{Decoder} &
    \textbf{\shortstack[c]{Decoder-\\ RI}} &
    \textbf{\shortstack[c]{Network-\\ RI}} \\
    \midrule
    IoU    & 0.887 & 0.884 & 0.922 & 0.927 & 0.927 & 0.929 \\
    NC     & 0.937 & 0.933 & 0.954 & 0.966 & 0.967 & 0.968 \\
    Epochs & 0     & 10    & 15    & 30    & 350   & 800   \\
    \bottomrule
  \end{tabular*}
\end{table}

\section{Conclusion and Limitations}
We introduced NoiseSDF2NoiseSDF, a framework that recovers clean surfaces from noisy, sparse point clouds using a Noise2Noise denoising strategy. Across diverse noise types, target generators, and denoising backbones, our method produces cleaner and smoother surfaces than prior baselines, both quantitatively and visually. Although our data-driven models require substantial training data and computation during training, our method can learn generalizable shape and noise priors that enable efficient denoising of unseen objects without further training or per-object optimization. However, the performance of our method may depend on the quality of the noisy supervision and the point-to-SDF target generator.
In the future, we aim to explore additional applications of NoiseSDF2NoiseSDF, such as scaling point cloud sizes for more complex geometries or replacing framework components with alternative architectures to improve noise representation and denoising performance.

\section*{Acknowledgment}
Shi Qiu is supported by The Chinese University of Hong Kong under Projects 4055212 and 6907743. The authors would like to thank Dr. Changkun Ye and Dr. Chamin Hewa Koneputugodage for insightful discussions.

\section*{Impact Statement}

This paper advances methods for learning neural signed distance functions from noisy supervision, which is useful when clean 3D ground truth is difficult or even impossible to obtain. This capability can improve the robustness of 3D reconstruction in domains where acquired geometry is inherently noisy or incomplete, such as robotics, augmented reality, and cultural heritage preservation. At the same time, the method may increase the risk of producing misleadingly ``clean'' shapes, particularly in safety-critical settings such as medicine. We therefore encourage practitioners in such settings to validate outputs against independent measurements rather than relying on visual plausibility alone.

\nocite{langley00}
\bibliography{example_paper}

\begin{thebibliography}{59}
\providecommand{\natexlab}[1]{#1}
\providecommand{\url}[1]{\texttt{#1}}
\expandafter\ifx\csname urlstyle\endcsname\relax
  \providecommand{\doi}[1]{doi: #1}\else
  \providecommand{\doi}{doi: \begingroup \urlstyle{rm}\Url}\fi

\bibitem[Atzmon \& Lipman(2020)Atzmon and Lipman]{atzmon2020sal}
Atzmon, M. and Lipman, Y.
\newblock {S}{A}{L}: Sign agnostic learning of shapes from raw data.
\newblock In \emph{Proceedings of the IEEE/CVF conference on computer vision and pattern recognition}, pp.\  2565--2574, 2020.

\bibitem[Atzmon \& Lipman(2021)Atzmon and Lipman]{atzmon2021sald}
Atzmon, M. and Lipman, Y.
\newblock {S}{A}{L}{D}: {S}ign {A}gnostic {L}earning with {D}erivatives.
\newblock In \emph{International Conference on Learning Representations (ICLR)}, 2021.

\bibitem[Batson \& Royer(2019)Batson and Royer]{pmlr-noise2self}
Batson, J. and Royer, L.
\newblock {N}oise2{S}elf: Blind denoising by self-supervision.
\newblock In Chaudhuri, K. and Salakhutdinov, R. (eds.), \emph{Proceedings of the 36th International Conference on Machine Learning}, volume~97 of \emph{Proceedings of Machine Learning Research}, pp.\  524--533. PMLR, 09--15 Jun 2019.

\bibitem[Ben-Shabat et~al.(2022)Ben-Shabat, Koneputugodage, and Gould]{ben2022digs}
Ben-Shabat, Y., Koneputugodage, C.~H., and Gould, S.
\newblock {D}i{G}{S}: {D}ivergence {G}uided {S}hape {I}mplicit {N}eural {R}epresentation for {U}noriented {P}oint {C}louds.
\newblock In \emph{IEEE/CVF Conference on Computer Vision and Pattern Recognition (CVPR)}, pp.\  19301--19310. IEEE, 2022.

\bibitem[Bora et~al.(2018)Bora, Price, and Dimakis]{bora2018ambientgan}
Bora, A., Price, E., and Dimakis, A.~G.
\newblock {A}mbient{G}{A}{N}: Generative models from lossy measurements.
\newblock In \emph{International Conference on Learning Representations}, 2018.

\bibitem[Boulch \& Marlet(2022)Boulch and Marlet]{boulch2022poco}
Boulch, A. and Marlet, R.
\newblock {P}o{C}o: {P}oint {C}onvolution for {S}urface {R}econstruction.
\newblock In \emph{Proceedings of the IEEE/CVF Conference on Computer Vision and Pattern Recognition (CVPR)}, pp.\  6302--6314, 2022.

\bibitem[Chabra et~al.(2020)Chabra, Lenssen, Ilg, Schmidt, Straub, Lovegrove, and Newcombe]{chabra2020dls}
Chabra, R., Lenssen, J.~E., Ilg, E., Schmidt, T., Straub, J., Lovegrove, S., and Newcombe, R.
\newblock {D}eep {L}ocal {S}hapes: {L}earning {L}ocal {S}{D}{F} {P}riors for {D}etailed {3D} {R}econstruction.
\newblock In \emph{Computer Vision -- ECCV 2020: 16th European Conference, Glasgow, UK, August 23--28, 2020, Proceedings, Part XXIX}, pp.\  608--625. Springer, 2020.

\bibitem[Chang et~al.(2015)Chang, Funkhouser, Guibas, Hanrahan, Huang, Li, Savarese, Savva, Song, Su, et~al.]{chang2015shapenet}
Chang, A.~X., Funkhouser, T., Guibas, L., Hanrahan, P., Huang, Q., Li, Z., Savarese, S., Savva, M., Song, S., Su, H., et~al.
\newblock {S}hape{N}et: An information-rich {3D} model repository.
\newblock \emph{arXiv preprint arXiv:1512.03012}, 2015.

\bibitem[Chen et~al.(2022)Chen, Liu, and Han]{chen2022lpi}
Chen, C., Liu, Y.-S., and Han, Z.
\newblock {L}atent {P}artition {I}mplicit with {S}urface {C}odes for {3D} {R}epresentation.
\newblock In \emph{European Conference on Computer Vision (ECCV) 2022}, pp.\  322--343. Springer, 2022.

\bibitem[Chen et~al.(2024)Chen, Han, and Liu]{chen2024localn2nm}
Chen, C., Han, Z., and Liu, Y.-S.
\newblock {I}nferring {N}eural {S}igned {D}istance {F}unctions by {O}verfitting on {S}ingle {N}oisy {P}oint {C}louds through {F}inetuning {D}ata-{D}riven based {P}riors.
\newblock In \emph{Advances in Neural Information Processing Systems (NeurIPS)}, 2024.

\bibitem[Chen \& Zhang(2019)Chen and Zhang]{chen2019imnet}
Chen, Z. and Zhang, H.
\newblock Learning implicit fields for generative shape modeling.
\newblock In \emph{Proceedings of the IEEE/CVF conference on computer vision and pattern recognition}, pp.\  5939--5948, 2019.

\bibitem[Chibane et~al.(2020)Chibane, Alldieck, and Pons-Moll]{chibane2020ifnet}
Chibane, J., Alldieck, T., and Pons-Moll, G.
\newblock {I}mplicit functions in feature space for {3D} shape reconstruction and completion.
\newblock In \emph{Proceedings of the IEEE/CVF conference on computer vision and pattern recognition}, pp.\  6970--6981, 2020.

\bibitem[Cui et~al.(2024)Cui, Qiu, Liu, Anwar, and Barnes]{cui2024numgrad}
Cui, R., Qiu, S., Liu, J., Anwar, S., and Barnes, N.
\newblock {N}um{G}rad-{P}ull: {N}umerical {G}radient {G}uided {T}ri-{P}lane {R}epresentation for {S}urface {R}econstruction from {P}oint {C}louds.
\newblock \emph{arXiv preprint arXiv:2411.17392}, 2024.

\bibitem[Dai et~al.(2020)Dai, Diller, and Nießner]{dai2020sgnn}
Dai, A., Diller, C., and Nießner, M.
\newblock {S}{G}-{N}{N}: {S}parse {G}enerative {N}eural {N}etworks for {S}elf-{S}upervised {S}cene {C}ompletion of {R}{G}{B}-{D} {S}cans.
\newblock In \emph{Proceedings of the IEEE/CVF Conference on Computer Vision and Pattern Recognition (CVPR)}, pp.\  849--858, 2020.

\bibitem[de~Silva~Edirimuni et~al.(2023)de~Silva~Edirimuni, Lu, Shao, Li, Robles-Kelly, and He]{de_Silva_Edirimuni_iterativepfn}
de~Silva~Edirimuni, D., Lu, X., Shao, Z., Li, G., Robles-Kelly, A., and He, Y.
\newblock {I}terative{P}{F}{N}: {T}rue {I}terative {P}oint {C}loud {F}iltering.
\newblock In \emph{Proceedings of the IEEE/CVF Conference on Computer Vision and Pattern Recognition (CVPR)}, pp.\  13530--13539, June 2023.

\bibitem[Du et~al.(2023)Du, Yu, Hussain, Armin, Petersson, and Li]{du2023weakly}
Du, H., Yu, X., Hussain, F., Armin, M.~A., Petersson, L., and Li, W.
\newblock Weakly-supervised point cloud instance segmentation with geometric priors.
\newblock In \emph{Proceedings of the ieee/cvf winter conference on applications of computer vision}, pp.\  4271--4280, 2023.

\bibitem[Erler et~al.(2020)Erler, Guerrero, Ohrhallinger, Mitra, and Wimmer]{erler2020points2surf}
Erler, P., Guerrero, P., Ohrhallinger, S., Mitra, N.~J., and Wimmer, M.
\newblock {P}oints2{S}urf: {L}earning {I}mplicit {S}urfaces from {P}oint {C}louds.
\newblock In \emph{Computer Vision -- ECCV 2020: 16th European Conference, Glasgow, UK, August 23--28, 2020, Proceedings, Part V}, volume 12350 of \emph{Lecture Notes in Computer Science}, pp.\  108--124. Springer, 2020.

\bibitem[Erler et~al.(2024)Erler, Fuentes-Perez, Hermosilla, Guerrero, Pajarola, and Wimmer]{erler2024ppsurf}
Erler, P., Fuentes-Perez, L., Hermosilla, P., Guerrero, P., Pajarola, R., and Wimmer, M.
\newblock {P}{P}{S}urf: {C}ombining {P}atches and {P}oint {C}onvolutions for {D}etailed {S}urface {R}econstruction.
\newblock In \emph{Computer Graphics Forum}, volume~43, pp.\  e15000. Wiley, 2024.

\bibitem[Genova et~al.(2020)Genova, Cole, Sud, Sarna, and Funkhouser]{genova2020localdeep}
Genova, K., Cole, F., Sud, A., Sarna, A., and Funkhouser, T.
\newblock {L}ocal deep implicit functions for {3D} shape.
\newblock In \emph{Proceedings of the IEEE/CVF conference on computer vision and pattern recognition}, pp.\  4857--4866, 2020.

\bibitem[Gropp et~al.(2020)Gropp, Yariv, Haim, Atzmon, and Lipman]{gropp2020igr}
Gropp, A., Yariv, L., Haim, N., Atzmon, M., and Lipman, Y.
\newblock {I}mplicit {G}eometric {R}egularization for {L}earning {S}hapes.
\newblock In \emph{Proceedings of the 37th International Conference on Machine Learning, ICML 2020, 13--18 July 2020, Virtual Event}, volume 119 of \emph{Proceedings of Machine Learning Research}, pp.\  3789--3799. PMLR, 2020.

\bibitem[Han et~al.(2018)Han, Yao, Yu, Niu, Xu, Hu, Tsang, and Sugiyama]{han2018co}
Han, B., Yao, Q., Yu, X., Niu, G., Xu, M., Hu, W., Tsang, I., and Sugiyama, M.
\newblock {C}o-teaching: {R}obust training of deep neural networks with extremely noisy labels.
\newblock \emph{Advances in neural information processing systems}, 31, 2018.

\bibitem[Hermosilla et~al.(2019)Hermosilla, Ritschel, and Ropinski]{hermosilla2019total}
Hermosilla, P., Ritschel, T., and Ropinski, T.
\newblock {T}otal denoising: Unsupervised learning of {3D} point cloud cleaning.
\newblock In \emph{Proceedings of the IEEE/CVF international conference on computer vision}, pp.\  52--60, 2019.

\bibitem[Hong et~al.(2022)Hong, Qiu, Li, Anwar, Harandi, Barnes, and Petersson]{hong2022pointcam}
Hong, J., Qiu, S., Li, W., Anwar, S., Harandi, M., Barnes, N., and Petersson, L.
\newblock Pointcam: Cut-and-mix for open-set point cloud learning.
\newblock \emph{arXiv preprint arXiv:2212.02011}, 2022.

\bibitem[Huang et~al.(2023)Huang, Gojcic, Atzmon, Litany, Fidler, and Williams]{huang2023nksr}
Huang, J., Gojcic, Z., Atzmon, M., Litany, O., Fidler, S., and Williams, F.
\newblock {N}eural kernel surface reconstruction.
\newblock In \emph{Proceedings of the IEEE/CVF Conference on Computer Vision and Pattern Recognition}, pp.\  4369--4379, 2023.

\bibitem[Huang et~al.(2021)Huang, Li, Jia, Lu, and Liu]{huang2021neighbor2neighbor}
Huang, T., Li, S., Jia, X., Lu, H., and Liu, J.
\newblock {N}eighbor2{N}eighbor: {S}elf-{S}upervised {D}enoising {F}rom {S}ingle {N}oisy {I}mages.
\newblock In \emph{Proceedings of the IEEE/CVF Conference on Computer Vision and Pattern Recognition (CVPR)}, pp.\  14781--14790, 2021.
\newblock \doi{10.1109/CVPR46437.2021.01454}.

\bibitem[Jiang et~al.(2021)Jiang, Cai, Zheng, and Xiao]{jiang2021neighborhood}
Jiang, H., Cai, J., Zheng, J., and Xiao, J.
\newblock {N}eighborhood-based neural implicit reconstruction from point clouds.
\newblock In \emph{2021 International Conference on 3D Vision (3DV)}, pp.\  1259--1268. IEEE, 2021.

\bibitem[Koch et~al.(2019)Koch, Matveev, Jiang, Williams, Artemov, Burnaev, Alexa, Zorin, and Panozzo]{koch2019abc}
Koch, S., Matveev, A., Jiang, Z., Williams, F., Artemov, A., Burnaev, E., Alexa, M., Zorin, D., and Panozzo, D.
\newblock {A}{B}{C}: A big cad model dataset for geometric deep learning.
\newblock In \emph{Proceedings of the IEEE/CVF conference on computer vision and pattern recognition}, pp.\  9601--9611, 2019.

\bibitem[Krull et~al.(2019)Krull, Buchholz, and Jug]{krull2019noise2void}
Krull, A., Buchholz, T.-O., and Jug, F.
\newblock {N}oise2{V}oid-learning denoising from single noisy images.
\newblock In \emph{Proceedings of the IEEE/CVF conference on computer vision and pattern recognition}, pp.\  2129--2137, 2019.

\bibitem[Langley(2000)]{langley00}
Langley, P.
\newblock Crafting papers on machine learning.
\newblock In Langley, P. (ed.), \emph{Proceedings of the 17th International Conference on Machine Learning (ICML 2000)}, pp.\  1207--1216, Stanford, CA, 2000. Morgan Kaufmann.

\bibitem[Lehtinen et~al.(2018)Lehtinen, Munkberg, Hasselgren, Laine, Karras, Aittala, and Aila]{pmlr-v80-lehtinen18a}
Lehtinen, J., Munkberg, J., Hasselgren, J., Laine, S., Karras, T., Aittala, M., and Aila, T.
\newblock {N}oise2{N}oise: {L}earning {I}mage {R}estoration without {C}lean {D}ata.
\newblock In \emph{Proceedings of the 35th International Conference on Machine Learning}, Proceedings of Machine Learning Research, pp.\  2965--2974, 2018.

\bibitem[Li et~al.(2024)Li, Gao, Liu, Gu, and Liu]{li2024implicitfiltering}
Li, S., Gao, G., Liu, Y., Gu, M., and Liu, Y.-S.
\newblock {I}mplicit filtering for learning neural signed distance functions from {3D} point clouds.
\newblock In \emph{European Conference on Computer Vision}, pp.\  234--251. Springer, 2024.

\bibitem[Lin et~al.(2022)Lin, Xiao, Shi, and Wang]{lin2022pgr}
Lin, S., Xiao, D., Shi, Z., and Wang, B.
\newblock {S}urface {R}econstruction from {P}oint {C}louds without {N}ormals by {P}arametrizing the {G}auss {F}ormula.
\newblock \emph{ACM Transactions on Graphics}, 42\penalty0 (2):\penalty0 1--19, 2022.

\bibitem[Lorensen \& Cline(1987)Lorensen and Cline]{lorensen1987marching}
Lorensen, W.~E. and Cline, H.~E.
\newblock {M}arching {C}ubes: A high resolution {3D} surface construction algorithm.
\newblock In \emph{Proceedings of the 14th annual conference on Computer graphics and interactive techniques}, pp.\  163--169. ACM, 1987.

\bibitem[Loshchilov \& Hutter(2019)Loshchilov and Hutter]{loshchilov2019decoupled}
Loshchilov, I. and Hutter, F.
\newblock {D}ecoupled {W}eight {D}ecay {R}egularization.
\newblock In \emph{International Conference on Learning Representations}, 2019.

\bibitem[Ma et~al.(2021)Ma, Han, Liu, and Zwicker]{ma2020neuralpull}
Ma, B., Han, Z., Liu, Y.-S., and Zwicker, M.
\newblock {N}eural-{P}ull: {L}earning {S}igned {D}istance {F}unctions from {P}oint {C}louds by {L}earning to {P}ull {S}pace onto {S}urfaces.
\newblock In \emph{International Conference on Machine Learning (ICML)}, 2021.

\bibitem[Ma et~al.(2023)Ma, Liu, and Han]{Ma2023SDF}
Ma, B., Liu, Y.-S., and Han, Z.
\newblock Learning {S}igned {D}istance {F}unctions from noisy {3D} point clouds via {N}oise to {N}oise {M}apping.
\newblock In \emph{Proceedings of the 40th International Conference on Machine Learning (ICML)}, volume 202 of \emph{Proceedings of Machine Learning Research}, pp.\  23338--23357. PMLR, 2023.

\bibitem[Mescheder et~al.(2019)Mescheder, Oechsle, Niemeyer, Nowozin, and Geiger]{mescheder2019occupancy}
Mescheder, L., Oechsle, M., Niemeyer, M., Nowozin, S., and Geiger, A.
\newblock {O}ccupancy networks: Learning {3D} reconstruction in function space.
\newblock In \emph{Proceedings of the IEEE/CVF conference on computer vision and pattern recognition}, pp.\  4460--4470, 2019.

\bibitem[Mi et~al.(2020)Mi, Luo, and Tao]{mi2020ssrnet}
Mi, Z., Luo, Y., and Tao, W.
\newblock Ssrnet: Scalable 3d surface reconstruction network.
\newblock In \emph{Proceedings of the IEEE/CVF conference on computer vision and pattern recognition}, pp.\  970--979, 2020.

\bibitem[Moran et~al.(2020)Moran, Schmidt, Zhong, and Coady]{moran2020noisier2noise}
Moran, N., Schmidt, D., Zhong, Y., and Coady, P.
\newblock {N}oisier2{N}oise: {L}earning to denoise from unpaired noisy data.
\newblock In \emph{Proceedings of the IEEE/CVF conference on computer vision and pattern recognition}, pp.\  12064--12072, 2020.

\bibitem[Park et~al.(2019)Park, Florence, Straub, Newcombe, and Lovegrove]{park2019deepsdf}
Park, J.~J., Florence, P., Straub, J., Newcombe, R., and Lovegrove, S.
\newblock {D}eep{S}{D}{F}: {L}earning continuous signed distance functions for shape representation.
\newblock In \emph{Proceedings of the IEEE/CVF conference on computer vision and pattern recognition}, pp.\  165--174, 2019.

\bibitem[Peng et~al.(2020)Peng, Niemeyer, Mescheder, Pollefeys, and Geiger]{peng2020convocc}
Peng, S., Niemeyer, M., Mescheder, L., Pollefeys, M., and Geiger, A.
\newblock {C}onvolutional {O}ccupancy {N}etworks.
\newblock In \emph{Computer Vision -- ECCV 2020: 16th European Conference, Glasgow, UK, August 23--28, 2020, Proceedings, Part III}, pp.\  523--540. Springer, 2020.

\bibitem[Peng et~al.(2021)Peng, Jiang, Liao, Niemeyer, Pollefeys, and Geiger]{peng2021sap}
Peng, S., Jiang, C., Liao, Y., Niemeyer, M., Pollefeys, M., and Geiger, A.
\newblock {S}hape {A}s {P}oints: {A} {D}ifferentiable {P}oisson {S}olver.
\newblock In \emph{Advances in Neural Information Processing Systems 34 (NeurIPS 2021)}, pp.\  13032--13044, 2021.

\bibitem[Qi et~al.(2017)Qi, Su, Mo, and Guibas]{qi2017pointnet}
Qi, C.~R., Su, H., Mo, K., and Guibas, L.~J.
\newblock {P}oint{N}et: Deep learning on point sets for {3D} classification and segmentation.
\newblock In \emph{Proceedings of the IEEE conference on computer vision and pattern recognition}, pp.\  652--660, 2017.

\bibitem[Quan et~al.(2020)Quan, Chen, Pang, and Ji]{quan2020self2self}
Quan, Y., Chen, M., Pang, T., and Ji, H.
\newblock {S}elf2{S}elf with dropout: {L}earning {S}elf-{S}upervised denoising from single image.
\newblock In \emph{Proceedings of the IEEE/CVF conference on computer vision and pattern recognition}, pp.\  1890--1898, 2020.

\bibitem[Tretschk et~al.(2020)Tretschk, Tewari, Golyanik, Zollh{\"o}fer, Stoll, and Theobalt]{tretschk2020patchnets}
Tretschk, E., Tewari, A., Golyanik, V., Zollh{\"o}fer, M., Stoll, C., and Theobalt, C.
\newblock {P}atch{N}ets: Patch-based generalizable deep implicit {3D} shape representations.
\newblock In \emph{Computer Vision--ECCV 2020: 16th European Conference, Glasgow, UK, August 23--28, 2020, Proceedings, Part XVI 16}, pp.\  293--309. Springer, 2020.

\bibitem[Wang et~al.(2024)Wang, Liu, Zhou, Wei, Deng, Murshed, and Lu]{wang2024noise4denoise}
Wang, W., Liu, X., Zhou, H., Wei, L., Deng, Z., Murshed, M.~M., and Lu, X.
\newblock {N}oise4{D}enoise: {L}everaging {N}oise for {U}nsupervised {P}oint {C}loud {D}enoising.
\newblock \emph{Computational Visual Media}, 10\penalty0 (4):\penalty0 659--669, 2024.

\bibitem[Wang et~al.(2023{\natexlab{a}})Wang, Wang, Wang, Dong, Gao, Chen, Xin, Tu, and Wang]{wang2023neuralimls}
Wang, Z., Wang, P., Wang, P., Dong, Q., Gao, J., Chen, S., Xin, S., Tu, C., and Wang, W.
\newblock {N}eural-{I}{M}{L}{S}: {S}elf-supervised implicit moving least-squares network for surface reconstruction.
\newblock \emph{IEEE Transactions on Visualization and Computer Graphics}, 2023{\natexlab{a}}.

\bibitem[Wang et~al.(2023{\natexlab{b}})Wang, Zhang, Xu, Zhang, Wang, Chen, Xin, Wang, and Tu]{zixiong23neuralsingular}
Wang, Z., Zhang, Y., Xu, R., Zhang, F., Wang, P.-S., Chen, S., Xin, S., Wang, W., and Tu, C.
\newblock {N}eural-{S}ingular-{H}essian: {I}mplicit {N}eural {R}epresentation of {U}noriented {P}oint {C}louds by {E}nforcing {S}ingular {H}essian.
\newblock \emph{ACM Transactions on Graphics (TOG)}, 42\penalty0 (6), 2023{\natexlab{b}}.

\bibitem[Wei et~al.(2025)Wei, Wang, Xu, Zhu, Sun, Li, Li, and Qin]{wei2025noise2score3d}
Wei, X., Wang, Y., Xu, A., Zhu, L., Sun, D., Li, K., Li, Y., and Qin, Q.
\newblock {N}oise2{S}core3{D}: {T}weedie's {A}pproach for {U}nsupervised {P}oint {C}loud {D}enoising.
\newblock In \emph{Proceedings of the IEEE/CVF International Conference on Computer Vision}, pp.\  25993--26003, 2025.

\bibitem[Xie et~al.(2020)Xie, Wang, and Ji]{xie2020noise2same}
Xie, Y., Wang, Z., and Ji, S.
\newblock {N}oise2{S}ame: Optimizing a {S}elf-{S}upervised bound for image denoising.
\newblock In \emph{Advances in Neural Information Processing Systems (NeurIPS)}, volume~33, 2020.

\bibitem[Yan et~al.(2022)Yan, Lin, Mitra, Lischinski, Cohen-Or, and Huang]{yan2022shapeformer}
Yan, X., Lin, L., Mitra, N.~J., Lischinski, D., Cohen-Or, D., and Huang, H.
\newblock {S}hape{F}ormer: {T}ransformer-based shape completion via sparse representation.
\newblock In \emph{Proceedings of the IEEE/CVF Conference on Computer Vision and Pattern Recognition}, pp.\  6239--6249, 2022.

\bibitem[Zhang \& Wonka(2025)Zhang and Wonka]{zhang2025lagem}
Zhang, B. and Wonka, P.
\newblock {L}a{G}e{M}: {A} {L}arge {G}eometry {M}odel for {3D} {R}epresentation {L}earning and {D}iffusion.
\newblock In \emph{The Thirteenth International Conference on Learning Representations}, 2025.

\bibitem[Zhang et~al.(2022)Zhang, Nießner, and Wonka]{zhang20223dilg}
Zhang, B., Nießner, M., and Wonka, P.
\newblock {3D}{I}{L}{G}: {I}rregular {L}atent {G}rids for {3D} {G}enerative {M}odeling.
\newblock In \emph{Advances in Neural Information Processing Systems 35 (NeurIPS 2022)}, 2022.

\bibitem[Zhang et~al.(2023)Zhang, Tang, Niessner, and Wonka]{zhang20233dshape2vecset}
Zhang, B., Tang, J., Niessner, M., and Wonka, P.
\newblock {3D}{S}hape2{V}ec{S}et: A {3D} shape representation for neural fields and generative diffusion models.
\newblock \emph{ACM Transactions On Graphics (TOG)}, 42\penalty0 (4):\penalty0 1--16, 2023.

\bibitem[Zhang \& Sabuncu(2018)Zhang and Sabuncu]{zhang2018generalized}
Zhang, Z. and Sabuncu, M.
\newblock {G}eneralized cross entropy loss for training deep neural networks with noisy labels.
\newblock \emph{Advances in neural information processing systems}, 31, 2018.

\bibitem[Zhao et~al.(2021)Zhao, Lei, Wen, Zhang, and Jia]{zhao2021signsal}
Zhao, W., Lei, J., Wen, Y., Zhang, J., and Jia, K.
\newblock {S}ign-agnostic implicit learning of surface self-similarities for shape modeling and reconstruction from raw point clouds.
\newblock In \emph{Proceedings of the IEEE/CVF Conference on Computer Vision and Pattern Recognition}, pp.\  10256--10265, 2021.

\bibitem[Zhou et~al.(2024)Zhou, Ma, Liu, and Han]{zhou2024fast}
Zhou, J., Ma, B., Liu, Y.-S., and Han, Z.
\newblock {F}ast learning of {S}igned {D}istance {F}unctions from {N}oisy {P}oint {C}louds via {N}oise to {N}oise {M}apping.
\newblock \emph{IEEE transactions on pattern analysis and machine intelligence}, 46\penalty0 (12):\penalty0 8936--8953, 2024.

\bibitem[Zhu et~al.(2017)Zhu, Park, Isola, and Efros]{zhu2017unpaired}
Zhu, J.-Y., Park, T., Isola, P., and Efros, A.~A.
\newblock {U}npaired image-to-image translation using cycle-consistent adversarial networks.
\newblock In \emph{Proceedings of the IEEE international conference on computer vision}, pp.\  2223--2232, 2017.

\bibitem[Zhu et~al.(2024)Zhu, Kang, Hui, Qian, Qiu, Dong, Bao, Heng, and Fu]{zhu2024ssp}
Zhu, R., Kang, D., Hui, K.-H., Qian, Y., Qiu, S., Dong, Z., Bao, L., Heng, P.-A., and Fu, C.-W.
\newblock {S}{S}{P}: Semi-signed prioritized neural fitting for surface reconstruction from unoriented point clouds.
\newblock In \emph{Proceedings of the IEEE/CVF Winter Conference on Applications of Computer Vision}, pp.\  3769--3778, 2024.

\end{thebibliography}
\bibliographystyle{icml2026}

\newpage
\appendix
\onecolumn
\section{Appendix}
\renewcommand{\thesection}{Appendix \Alph{section}}
\renewcommand{\thesubsection}{\thesection.\arabic{subsection}}

\counterwithin{figure}{section}
\counterwithin{table}{section}
\counterwithin{equation}{section}

\subsection{Metrics Formula}
We detail the evaluation metrics adopted in our experiments. 

\[
\begin{aligned}
\mathbf{CD}(P,Q)
&=\frac{1}{|P|}\sum_{p\in P}\min_{q\in Q}\|p-q\|_2
 \;+\;
 \frac{1}{|Q|}\sum_{q\in Q}\min_{p\in P}\|q-p\|_2,\\[1ex]
\mathbf{F}_1(\tau)
&=\frac{2\,\mathrm{precision}(\tau)\,\mathrm{recall}(\tau)}
       {\mathrm{precision}(\tau)+\mathrm{recall}(\tau)},
 \quad \tau=0.02,\\[1ex]
\mathbf{NC}
&=\frac{1}{N}\sum_{i=1}^N
 \bigl|\langle n_i^{\mathrm{pred}},\,n_i^{\mathrm{gt}}\rangle\bigr|,\\[1ex]
\mathbf{MNC}(M)
&=\frac{1}{|E|}\sum_{e=(v_0,v_1)\in E}
 \Biggl(1
 -\frac{\bigl\langle (v_1-v_0)\times(a_e-v_0),\,(b_e-v_0)\times(v_1-v_0)\bigr\rangle}
        {\|(v_1-v_0)\times(a_e-v_0)\|\;\|(b_e-v_0)\times(v_1-v_0)\|}\Biggr).
\end{aligned}
\]

Here \(P\) and \(Q\) are the sets of predicted and ground-truth 3D samples, with \(|\cdot|\) denoting cardinality. The Chamfer Distance (\(\mathbf{CD}\)) measures the average nearest-neighbor distance between the two sets in both directions. It is non-negative (\(\geq 0\)), and lower values indicate better geometric alignment.

The threshold \(\tau\) (set to 0.02) in the \(\mathbf{F}_1(\tau)\) score defines precision and recall by counting how many nearest-neighbor distances fall below \(\tau\). The F1 score, computed as the harmonic mean of precision and recall, ranges from 0 to 1, with higher values indicating better correspondence between predicted and ground-truth points.

In \(\mathrm{NC}\), \(n_i^{\mathrm{pred}}\) and \(n_i^{\mathrm{gt}}\) denote the predicted and ground-truth unit normals at sample \(i\) (out of \(N\) total samples). The Normal Consistency is the mean absolute dot product between matched normals, ranging from 0 to 1. Higher values imply better alignment of surface orientation.

\(\mathrm{MNC}(M)\) denotes Mesh Normal Consistency for a mesh \(M\) with edge set \(E\). Each edge \(e=(v_0,v_1)\) is shared by two faces whose opposite vertices are \(a_e\) and \(b_e\). The unnormalized face normals are computed via cross products \((v_1 - v_0) \times (a_e - v_0)\) and \((b_e - v_0) \times (v_1 - v_0)\), and their cosine similarity measures the local smoothness across the edge. The value of \(\mathrm{MNC}\) is averaged over all edges, and typically falls in the range \([0, 2]\), where lower values correspond to smoother and more consistent surface geometry.

\subsection{More Visualization }

\begin{figure}[!htbp]
    \centering
    \includegraphics[width=1\linewidth]{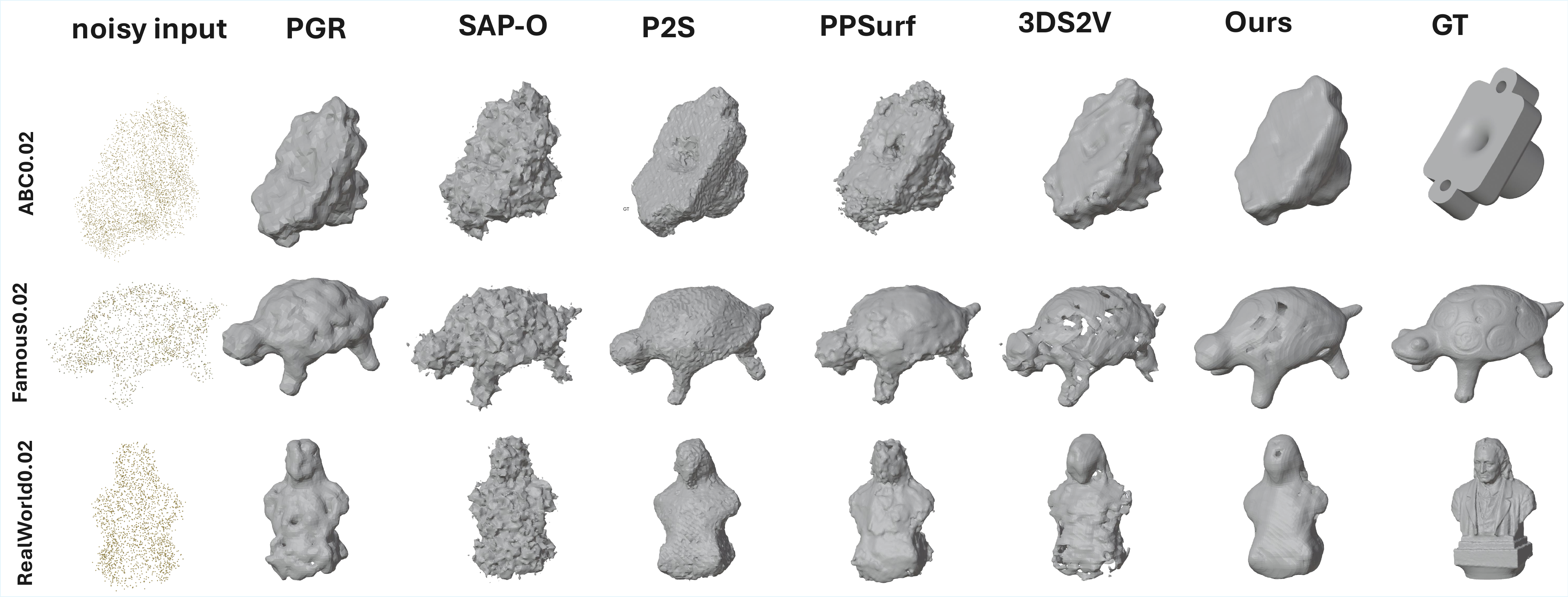}
    \caption{More results on ABC, Famous and Real under the noise level of 0.02}
\end{figure}

\begin{figure}[!htbp]
    \centering
    \includegraphics[width=0.8\linewidth]{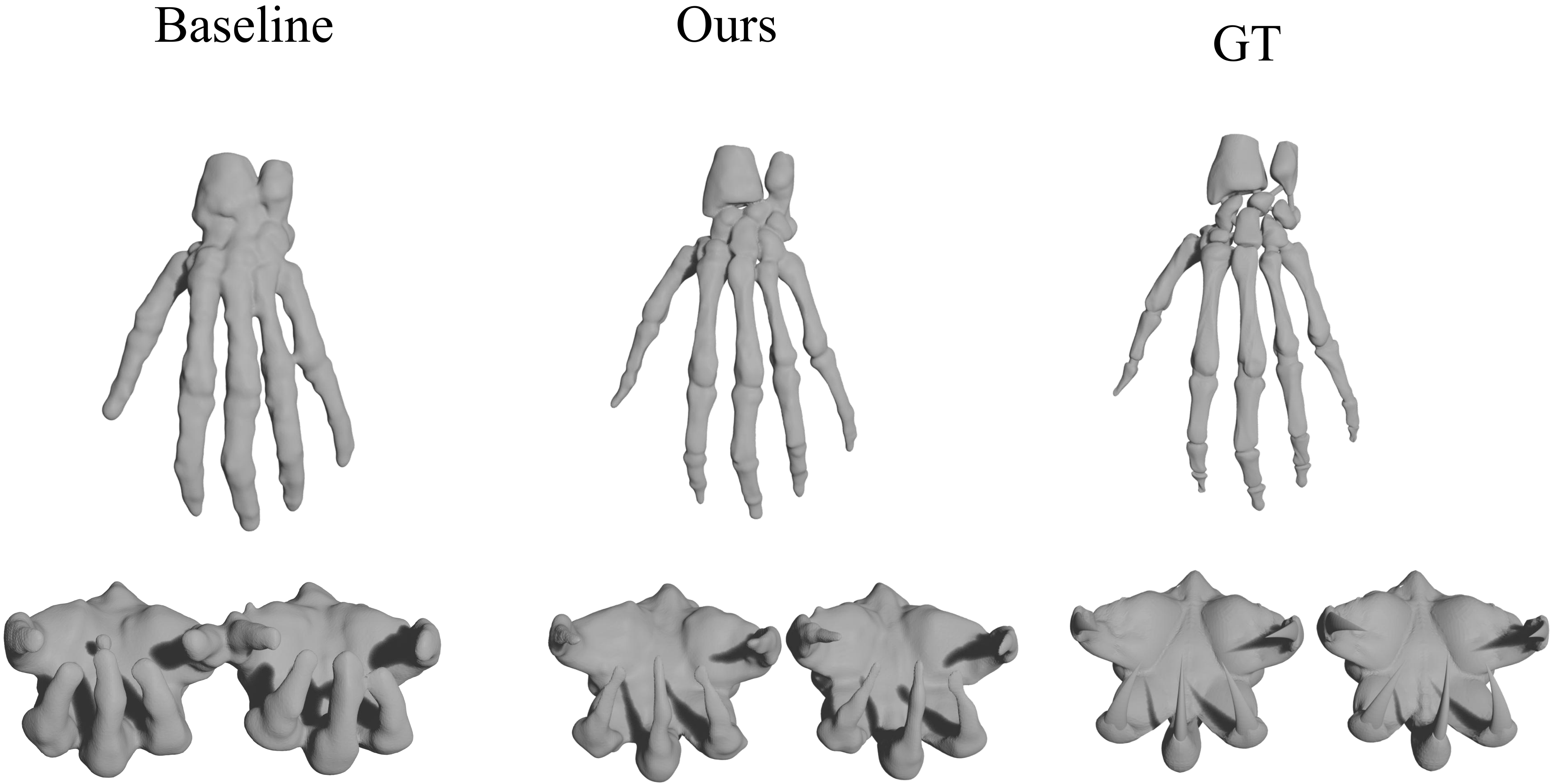}
    \caption{\textbf{Additional complex shape examples under 0.01 level Noise.} From left to right: LaGeM~\cite{zhang2025lagem}, Ours, and Ground Truth. These cases contain thin elongated parts and complex local geometry, including multi-branch junctions, narrow protrusions, and self-occluded regions. Compared with the noisy inputs, the denoised results exhibit smoother surfaces and more coherent local geometry; in particular, the articulated connecting structure becomes much clearer after denoising, while the claw-like tips are recovered with noticeably sharper ends, making the results visually much closer to the ground truth.
}
\end{figure}

\subsection{More Quantitative Results}

\begin{table*}[htbp]
\captionsetup{justification=raggedright, width=\textwidth, skip=5pt}
\caption{
Comparison of Neural-Singular-Hessian (NSH)~\citep{zixiong23neuralsingular}, the traditional method Poisson Reconstruction (Poisson), and NoiseSDF2NoiseSDF (Ours) across six noisy test datasets.
}
\label{tab:data_comparison}
\centering
\scriptsize
\setlength{\tabcolsep}{2pt}
\renewcommand{\arraystretch}{1.1}
\rowcolors{2}{blue!5}{white}
\begin{adjustbox}{max width=\textwidth}
\begin{tabular}{lcccccccccccccccc}
\toprule
\textbf{Dataset ($\sigma$)} & \multicolumn{3}{c}{\textbf{NC} \(\uparrow\)} & 
\multicolumn{3}{c}{\textbf{Mesh NC} \(\downarrow\)} & 
\multicolumn{3}{c}{\textbf{Chamfer} \(\downarrow\)} & 
\multicolumn{3}{c}{\textbf{F-Score} \(\uparrow\)} \\ 
\cmidrule(lr){2-4}\cmidrule(lr){5-7}\cmidrule(lr){8-10}\cmidrule(lr){11-13}
& NSH & Poisson & Ours & NSH & Poisson & Ours & NSH & Poisson & Ours & NSH & Poisson & Ours \\
\midrule

ABC (0.01)    
& 0.845  & 0.804  & \textbf{0.865}
& 0.026  & 0.074  & \textbf{0.024}
& 0.016  & 0.058  & \textbf{0.014}
& 0.933  & 0.771  & \textbf{0.938} \\

ABC (0.02)    
& 0.752  & 0.743  & \textbf{0.812}
& 0.019  & 0.091  & \textbf{0.018}
& \textbf{0.026} & 0.078  & 0.032
& \textbf{0.786} & 0.679  & 0.724 \\

Famous (0.01)
& 0.784  & 0.783  & \textbf{0.831}
& 0.028  & 0.072  & \textbf{0.025}
& 0.017  & 0.036  & \textbf{0.016}
& 0.941  & 0.847  & \textbf{0.941} \\

Famous (0.02)
& 0.706  & 0.728  & \textbf{0.767}
& 0.035  & 0.090  & \textbf{0.024}
& \textbf{0.027} & 0.049  & 0.034
& \textbf{0.760} & 0.757  & 0.726 \\

Real (0.01)
& 0.790  & 0.806  & \textbf{0.845}
& \textbf{0.030} & 0.065 & 0.031
& 0.016  & 0.052  & \textbf{0.015}
& 0.953  & 0.783  & \textbf{0.956} \\

Real (0.02)
& 0.713  & 0.721  & \textbf{0.793}
& 0.023  & 0.078  & \textbf{0.020}
& \textbf{0.025} & 0.087  & 0.026
& 0.783  & 0.643  & \textbf{0.809} \\

\textbf{mean (0.01)}
& 0.806 & 0.798 & \textbf{0.847}
& 0.028 & 0.070 & \textbf{0.027}
& 0.016 & 0.049 & \textbf{0.015}
& 0.942 & 0.800 & \textbf{0.945} \\

\textbf{mean (all)}
& 0.765 & 0.764 & \textbf{0.819}
& 0.027 & 0.078 & \textbf{0.024}
& \textbf{0.021} & 0.060 & 0.023
& \textbf{0.859} & 0.747 & 0.849 \\

\bottomrule
\end{tabular}
\end{adjustbox}
\end{table*}

\subsubsection{Denoising-Then-Reconstruction comparison}

To contrast with our end-to-end framework, we evaluated a \emph{two-stage denoising–then–reconstruction} pipeline, in which point cloud denoising and surface reconstruction are performed as separate modules. Specifically, we employed IterativePFN~\citep{de_Silva_Edirimuni_iterativepfn} for point cloud denoising and used 3DS2V as the subsequent Point2SDF surface reconstruction module. The evaluation was conducted on the ShapeNet Chair category under Gaussian noise of $\sigma = 0.01$, and the quantitative results in Table~\ref{tab:denoise_recon} show that our method achieves superior reconstruction quality.

\begin{table*}[htbp]
\captionsetup{justification=raggedright, width=\textwidth, skip=5pt}
\caption{
Comparison between a two-stage denoising--then--reconstruction pipeline and NoiseSDF2NoiseSDF (Ours).
}
\label{tab:denoise_recon}
\centering
\scriptsize
\setlength{\tabcolsep}{2pt}
\renewcommand{\arraystretch}{1.1}
\rowcolors{2}{blue!5}{white}
\begin{adjustbox}{max width=\textwidth}
\begin{tabular}{lcccc}
\toprule
\textbf{Method} & \textbf{IoU} $\uparrow$ & \textbf{NC} $\uparrow$ & \textbf{CD-L2} $\downarrow$ & \textbf{F-Score} $\uparrow$ \\
\midrule
Denoising-Then-Reconstruction & 0.882 & 0.954 & 0.016 & 0.969 \\
\textbf{Ours}                 & \textbf{0.927} & \textbf{0.966} & \textbf{0.013} & \textbf{0.986} \\
\bottomrule
\end{tabular}
\end{adjustbox}
\end{table*}

\subsubsection{Multiple noise-to-noise mappings}

We explored whether extending NoiseSDF2NoiseSDF to a multi-target Noise2Noise formulation can further stabilize training. In addition to the standard pairwise (1-to-1) setup, we implemented a \emph{1-to-3 Noise2Noise mapping}, where in each training step the Point2SDF network processes three independently corrupted point clouds, and the mean of their predicted SDF values is used as supervision. The experiment follows the ablation settings on the ShapeNet Chair dataset with Gaussian noise of $\sigma = 0.01$. The 1-to-3 configuration indeed produces a slightly more stable training process with lower loss, but it does not improve reconstruction quality: the final IoU, NC, Chamfer, and F-Score are essentially the same as in the 1-to-1 case, as summarized in Table~\ref{tab:multi_target}. 

\begin{table*}[htbp]
\captionsetup{justification=raggedright, width=\textwidth, skip=5pt}
\caption{
Quantitative comparison between the standard 1-to-1 Noise2Noise formulation and the 1-to-3 multi-target variant.
}
\label{tab:multi_target}
\centering
\scriptsize
\setlength{\tabcolsep}{2pt}
\renewcommand{\arraystretch}{1.1}
\rowcolors{2}{blue!5}{white}
\begin{adjustbox}{max width=\textwidth}
\begin{tabular}{lcccc}
\toprule
\textbf{Method} & \textbf{IoU} $\uparrow$ & \textbf{NC} $\uparrow$ & \textbf{Chamfer} $\downarrow$ & \textbf{F-Score} $\uparrow$ \\
\midrule
3DS2V  & 0.887 & 0.937 & 0.014 & 0.986 \\
1-to-1 & 0.927 & 0.966 & 0.013 & 0.986 \\
1-to-3 & 0.923 & 0.967 & 0.013 & 0.986 \\
\bottomrule
\end{tabular}
\end{adjustbox}
\end{table*}

\subsubsection{Realistic LiDAR Sensing Conditions}

To assess robustness under more realistic sensing conditions, we further evaluated our method using a LiDAR-style noise model. Unlike additive i.i.d. perturbations, real range sensors exhibit structured artifacts such as depth-dependent noise, spatially irregular sampling, dropout, and sporadic outlier returns. These effects introduce anisotropic and highly non-uniform corruption patterns that pose a greater challenge for surface reconstruction.

We considered three corruption levels, with higher levels introducing stronger depth noise, more aggressive dropout, and a larger proportion of outlier returns. The evaluation was conducted on the ShapeNet Chair dataset, where ground-truth meshes are available for quantitative assessment. The corresponding results are summarized in Table~\ref{tab:lidar_noise}, covering both Chamfer Distance (CD) and F1 under progressively severe LiDAR-style corruptions.

\begin{table*}[htbp]
\captionsetup{justification=raggedright, width=\textwidth, skip=5pt}
\caption{
Performance under LiDAR-style structured noise at three corruption levels on the ShapeNet Chair dataset.
}
\label{tab:lidar_noise}
\centering
\scriptsize
\setlength{\tabcolsep}{2pt}
\renewcommand{\arraystretch}{1.1}
\rowcolors{2}{blue!5}{white}
\begin{adjustbox}{max width=\textwidth}
\begin{tabular}{lcccc}
\toprule
\multirow{2}{*}{\textbf{Corruption Level}}
& \multicolumn{2}{c}{\textbf{CD} $\downarrow$}
& \multicolumn{2}{c}{\textbf{F1} $\uparrow$} \\
\cmidrule(lr){2-3} \cmidrule(lr){4-5}
& 3DS2V & Ours & 3DS2V & Ours \\
\midrule
\textbf{Low}  & 0.016 & \textbf{0.012} & 0.961 & \textbf{0.990} \\
\textbf{Mid}  & 0.031 & \textbf{0.023} & 0.772 & \textbf{0.878} \\
\textbf{High} & 0.059 & \textbf{0.052} & 0.518 & \textbf{0.601} \\
\bottomrule
\end{tabular}
\end{adjustbox}
\end{table*}

\subsection{Theoretical Justification}

We present a first-order analysis showing that perturbing the closest point with zero-mean Gaussian noise yields a noisy signed distance whose expectation, to first order, approximately matches the clean signed distance at the query point, including on the zero level set.

Let $\phi : \mathbb{R}^d \to \mathbb{R}$ be a signed distance function with zero level set $S = \{ x : \phi(x) = 0 \}$, satisfying the Eikonal equation $|\nabla \phi(x)| = 1$. For a query point $q$, let $p$ be its closest point on $S$, define the unit normal $n := \nabla \phi(p)$, and write
$$
q = p + s n, \quad s = \phi(q).
$$
We consider a perturbation of the closest point $p$ by zero-mean Gaussian noise, $\tilde p = p + \varepsilon$ with $\varepsilon \sim \mathcal{N}(0, \Sigma_p)$, and study the induced perturbation of the signed distance, $\delta s := \tilde s - s$. 

\noindent \textbf{Analysis}: We perform a first-order Taylor expansion around $p$:
$$
\phi(x) \approx \phi(p) + \nabla \phi(p)^\top (x - p) = n^\top (x - p),
$$
since $\phi(p) = 0$. Thus, near $p$ the signed distance is approximately the projection onto the normal. Approximating the signed distance with respect to the perturbed reference $\tilde p$ and treating $n$ as constant to first order, we obtain
$$
\tilde s \approx n^\top (q - \tilde p) = n^\top (q - (p + \varepsilon)) = n^\top (q - p) - n^\top \varepsilon \approx s - n^\top \varepsilon.
$$
Hence, 
$$
\delta s = \tilde s - s \approx -n^\top \varepsilon,
$$
which is a linear functional of $\varepsilon$. Since $\varepsilon$ is zero-mean Gaussian, we have
$$
\mathbb{E}[\delta s] \approx -n^\top \mathbb{E}[\varepsilon] = 0, \quad \text{and therefore} \quad \mathbb{E}[\tilde s(q)] \approx s(q).
$$
Crucially, this reasoning does \emph{not} invoke a condition of the form $|\varepsilon| \ll |s(q)|$ and remains valid at the zero-level set. If $q \in S$, then $s = \phi(q) = 0$ and $q = p$, so the first-order expression becomes
$$
\tilde s \approx -n^\top \varepsilon, \quad \delta s \approx -n^\top \varepsilon,
$$
with
$$
\mathbb{E}[\tilde s] \approx 0 = s.
$$
Overall, this normal-based analysis provides a first-order unbiasedness guarantee for the noisy SDF both away from and exactly at the zero-level set.

\subsection{Network Architecture}

\begin{figure}[!h]
    \centering
    \includegraphics[width=1\linewidth]{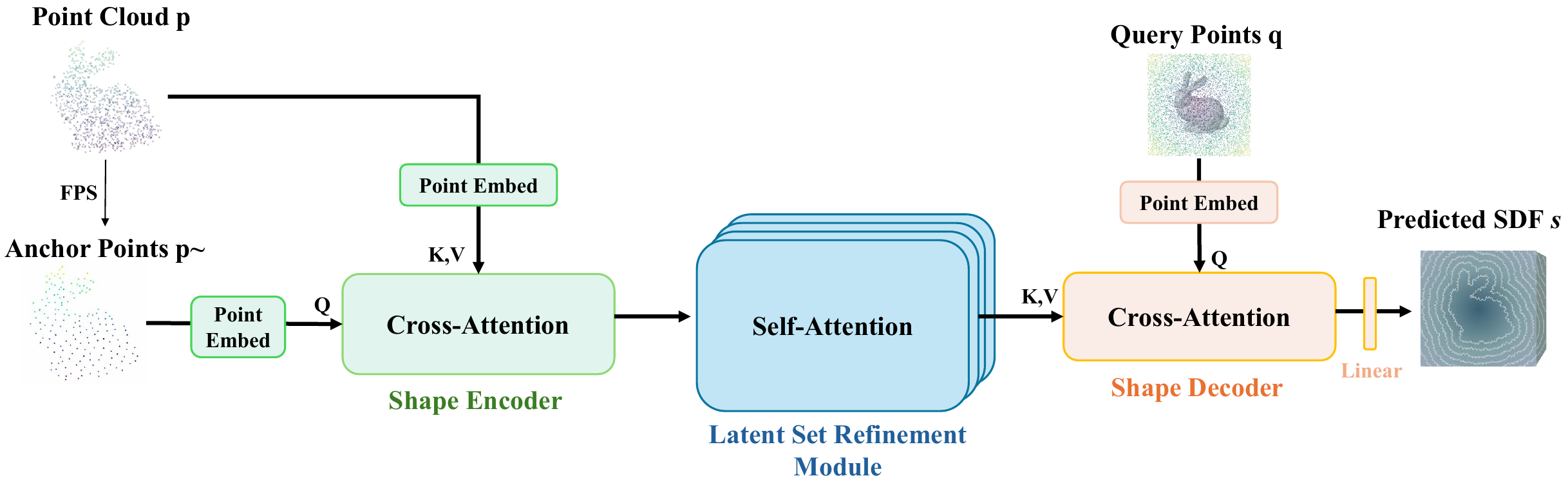}
    \caption{\textbf{Network architecture} of the proposed SDF predictor. The Shape Encoder compresses the
input point cloud into a set of latent tokens, the Latent Set Refinement Module refines
them via self-attention, and the Shape Decoder predicts the signed distance at each
query point.}
    \label{fig:net_architecture}
\end{figure}

\Cref{fig:net_architecture} illustrates the architecture of our SDF predictor,
which is adapted from 3DShape2VecSet~\cite{zhang20233dshape2vecset}.
Given a noisy input point cloud
$\mathbf{p}$ with $N=2048$ points, the Shape Encoder first applies farthest point
sampling (FPS) to obtain $M=512$ anchor points $\mathbf{p}{\sim}$. Both
$\mathbf{p}$ and $\mathbf{p}{\sim}$ are embedded by point embeddings. The
anchor features serve as queries $Q$, while the input point features serve as
keys and values $(K,V)$, allowing cross-attention to aggregate the noisy point
cloud into a set of 512 latent tokens. These tokens are further refined by the
Latent Set Refinement Module, which consists of 24 self-attention layers. Given
a query point $\mathbf{q}$, the Shape Decoder embeds $\mathbf{q}$ and uses it as
the query $Q$ to cross-attend to the refined latent tokens as $(K,V)$. A final
linear layer predicts the signed distance value $s$ at $\mathbf{q}$.

\end{document}